\documentclass[journal]{IEEEtran}

\usepackage{amsmath}
\usepackage{amssymb}
\usepackage{makecell}
\usepackage{ragged2e}
\usepackage{mathrsfs}
\usepackage{epsfig}
\usepackage{graphicx}
\usepackage{geometry}
\graphicspath{{Images/}}
\DeclareGraphicsExtensions{.pdf, .png, .jpg, .eps}

\usepackage{multirow} 
\usepackage{multicol} 
\usepackage[colorinlistoftodos]{todonotes}
\usepackage{array, textcomp, stackrel, url, mathtools, enumerate}

\usepackage{float}
\usepackage{subfig}
\usepackage{footnote}
\usepackage{tablefootnote}
\usepackage{color}
\usepackage{graphicx}
\usepackage{colortbl}
\usepackage{comment}
\usepackage{cite}
\usepackage{amsmath}
\usepackage{caption}
\usepackage{arydshln}
\usepackage[pagewise]{lineno}

\title{MobileXNet: An Efficient Convolutional Neural Network for Monocular Depth Estimation}

\author{Xingshuai Dong, Matthew A. Garratt,~\IEEEmembership{Senior Member,~IEEE}, Sreenatha G. Anavatti, and Hussein A. Abbass,~\IEEEmembership{Fellow,~IEEE} 
	
School of Engineering and Information Technology, University of New South Wales, Canberra, Australia.
}

\begin{document}

\maketitle
 
\begin{abstract}
Depth is a vital piece of information for autonomous vehicles to perceive obstacles. Due to the relatively low price and small size of monocular cameras, depth estimation from a single RGB image has attracted great interest in the research community. In recent years, the application of Deep Neural Networks (DNNs) has significantly boosted the accuracy of monocular depth estimation (MDE). State-of-the-art methods are usually designed on top of complex and extremely deep network architectures, which require more computational resources and cannot run in real-time without using high-end GPUs. Although some researchers tried to accelerate the running speed, the accuracy of depth estimation is degraded because the compressed model does not represent images well. In addition, the inherent characteristic of the feature extractor used by the existing approaches results in severe spatial information loss in the produced feature maps, which also impairs the accuracy of depth estimation on small sized images. In this study, we are motivated to design a novel and efficient Convolutional Neural Network (CNN) that assembles two shallow encoder-decoder style subnetworks in succession to address these problems. In particular, we place our emphasis on the trade-off between the accuracy and speed of MDE. Extensive experiments have been conducted on the NYU depth v2, KITTI, Make3D and Unreal data sets. Compared with the state-of-the-art approaches which have an extremely deep and complex architecture, the proposed network not only achieves comparable performance but also runs at a much faster speed on a single, less powerful GPU.
\end{abstract}



\section{Introduction}
Autonomous vehicles have been extensively used in many applications, such as aerial surveillance, search and rescue. In order to safely operate in cluttered and unpredictable environments, these vehicles require a strong awareness of their operational surroundings, in particular, the ability to detect and avoid stationary or mobile obstacles \cite{yang2019fast, eom2020temporally}. Depth estimation provides a geometry-independent paradigm in order to detect free, navigable space with the minimum safe distance. For the sake of obtaining depth information, most previous research depends on active sensors, such as LiDAR and the RGB-D cameras. However, these devices are bulky and energy consuming, which inhibit their deployment on limited payload platforms, for example, Micro Aerial Vehicles (MAVs). In contrast, monocular RGB cameras are lightweight and have low power consumption. More importantly, they provide richer information about the environment, which can be processed and applied in real-time. As a result, depth estimation from single RGB images has become a desirable alternative to active sensors. \par
\begin{figure}[!t]
    \centering
    \includegraphics[width = .9\linewidth]{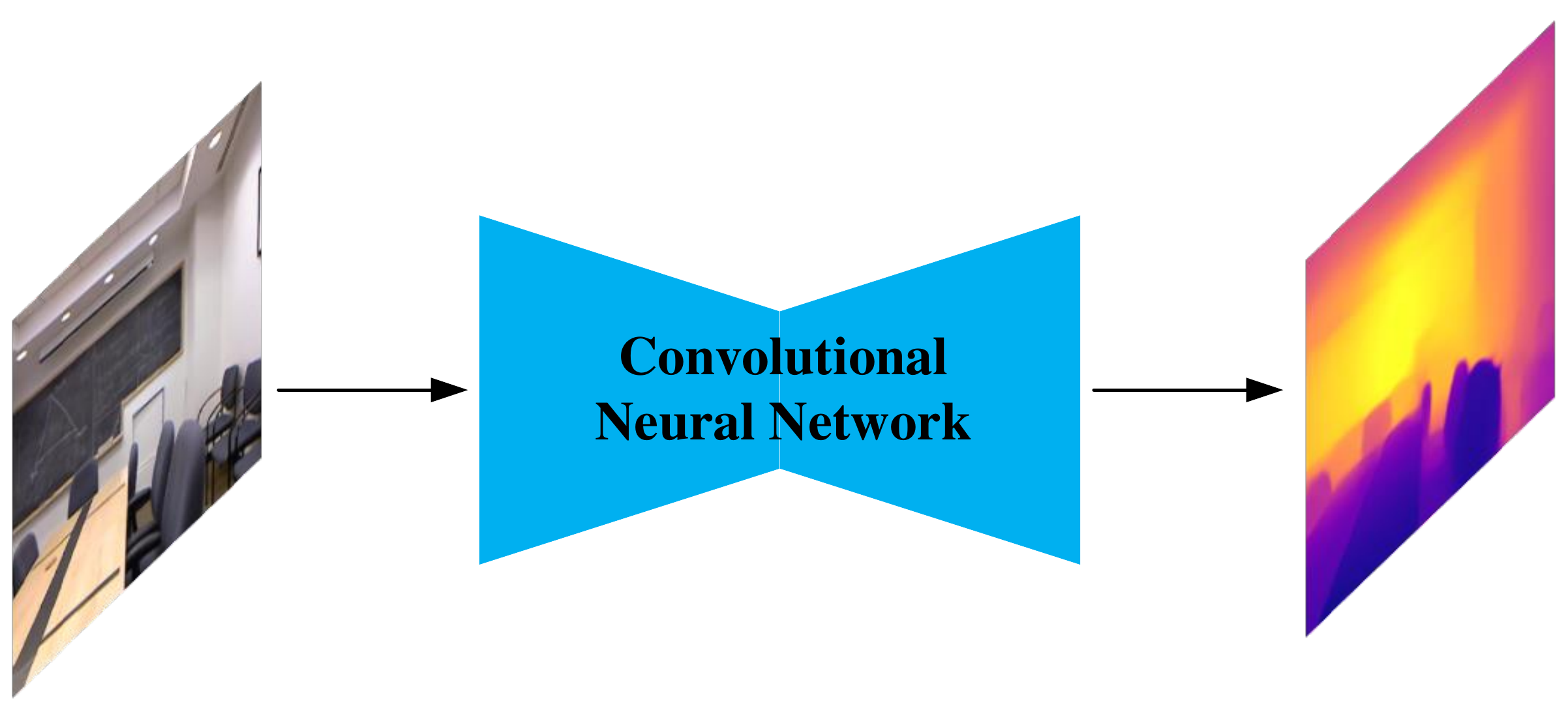} \\
    \caption{The overall pipeline of CNN-based monocular depth estimation (best viewed in color).}
    \label{fig:overallpipeline}
\end{figure}
State-of-the-art methods \cite{laina2016deeper, cao2017estimating, li2018monocular, xu2018structured, ma2018sparse, hu2019revisiting, chen2019structure, ye2020dpnet} normally utilize CNNs to learn features in order to predict a depth value for each pixel, and significantly outperform their classical counterparts \cite{saxena2006learning, saxena2008make3d, karsch2014depth}. However, these methods were developed based on extremely deep and complex network architectures. Therefore, they suffer a heavy computational cost and cannot be run in real-time without using high-end GPUs. Therefore, it is impractical to deploy those methods on platforms where reaction time is critical, such as small drones. \par
On the other hand, Wofk et al. \cite{wofk2019fastdepth} introduced an encoder-decoder network using the lightweight CNN \cite{howard2017mobilenets}. Network pruning techniques \cite{yang2018netadapt} were used to further reduce the size of the model. Despite their method achieving a significant improvement in speed, the accuracy was not comparable to the state-of-the-art results. \par 
In general, the above-mentioned methods employ the fully convolutional part of the CNNs \cite{he2016deep, huang2017densely, hu2018squeeze, howard2017mobilenets} designed for image classification as the feature extractor/encoder. However, these networks downsample the resolution of the final feature maps to $1/32$ scale of the input image. Although successive downsampling increases the receptive field and enables the generated feature maps to include more semantic information, it results in loss of spatial information. It is difficult to accurately recover this type of information from these feature maps. Hence, the accuracy of depth estimation will be impaired, in particular, when small sized images are processed. Moreover, the loss functions (i.e., the $L_{1}$ loss, $L_{2}$ loss and berHu loss) used by those methods are insensitive to the errors which occur at step edges. This leads to distorted and blurry edges in the estimated depth maps. \par
To address the above problems, we propose a novel CNN and a hybrid loss function for monocular depth estimation (MDE) in this study. Particularly, we aim to develop a network which has a proper trade-off between accuracy and speed. To be specific, the proposed network assembles two relatively shallow encoder-decoder style subnetworks back-to-back in a unified framework. Compared with the previous studies \cite{laina2016deeper, cao2017estimating, xu2018structured, li2018monocular, ma2018sparse, hu2019revisiting, chen2019structure, ye2020dpnet}, our network has a much shallower and simpler architecture as it is built on top of a series of simple convolutional layers rather than Inception modules \cite{szegedy2015going} or Residual blocks \cite{he2016deep}. In addition, each subnetwork involves less downsampling operations. This enables the intermediate feature maps to have a larger resolution and contain more spatial information, which is beneficial for depth estimation on small sized images. To the best of our knowledge, this work is the first attempt that stacks shallow encoder-decoder style CNNs in a unified framework to avoid the loss of spatial information and enhance the network's representative ability for depth estimation. Different from existing work \cite{li2018monocular, ye2020dpnet}, our method avoids the loss of spatial details but does not incorporate a separate spatial branch or fuse the hierarchical features of the encoder network. To preserve more edge details of objects, we design a hybrid loss function which adds constraints on the gradient data of depth maps. \par
Since the proposed network is designed for autonomous driving and mobile robots as well as is inspired by the XNet \cite{bullock2019xnet}, we refer to it as ``MobileXNet" in this paper. Compared with the XNet network, MobileXNet has three different characteristics. First, it is developed for the regression application rather than the classification task. In addition, the encoder of the first subnetwork consists of a series of depthwise separable convolutional layers \cite{chollet2017xception}, which enables a faster computational speed. Finally, a bridge module is inserted between the encoder and decoder in each subnetwork, including a set of dilated convolutional layers with different dilation rates, to capture the context information in multiple scales. This configuration generates better results than those derived using a single dilation rate. As a result, the proposed MobileXNet can be used for depth estimation in a faster and more effective manner in contrast to the XNet method. \par
The contributions of this study are summarized as follows: (1) We introduce a novel CNN architecture for achieving running efficient and accurate depth estimation from a single image; (2) We design a hybrid loss function; (3) We evaluate the proposed method on a dataset consisting of small sized synthetic images. In particular, we show that the size of the filters used in the input layer of CNNs has an influence on the performance of MDE when small sized images are processed; (4) We use Pareto Optimality to compare the error and running time over different methods, which has not been exploited in depth estimation; and (5) We demonstrate that the proposed method not only generates comparable accuracy to the state-of-the-art methods which use either extremely deep and complex architecture or post-processing but also runs much faster on a single less powerful GPU.\par
The rest of this paper is organized as follows. The related work is reviewed in Section II. In Section III, our methodology is introduced. The experimental setup and results are reported in Sections IV and V respectively. Finally, conclusions and future work are presented in Section VI.
\section{Related Work}
In this section, we review the related works for monocular depth estimation. We categorize the reviewed methods into traditional handcrafted feature-based methods and deep learning-based methods.  
\subsection{Traditional Handcrafted Feature-based Methods}
Monocular depth estimation (MDE) has been an active research topic in the research community of computer vision and autonomous driving. The early works mainly depend on hand-crafted features and probabilistic graphical models. Saxena et al. \cite{saxena2006learning} exploited the absolute and relative depth features and Markov Random Fields (MRF) to predict depth from monocular images. Later, the authors extended their work to 3D scene reconstruction \cite{saxena2008make3d}. Inspired by \cite{saxena2008make3d}, Liu et al. \cite{liu2010single} proposed a method which uses semantic labels as the contextual information for MDE. They use a learned MRF to infer the semantic class for each pixel, and then apply the L-BFGS algorithm to obtain a pixel-wise depth image.

In addition to the above mentioned parametric methods, MDE can also be solved as a non-parametric problem by inferring the whole depth map from candidate depth maps. Karsch et al. \cite{karsch2014depth} introduced a pixel transfer-based method. Given an input image, they first search similar images from a dataset by matching GIST features. A set of possible depth values for the scene are constructed by applying label transfer between the given input image and the matched image. Liu et al. \cite{liu2014discrete} designed a discrete-continuous conditional random field (CRF) to avoid the over-smoothing and maintain occlusion boundaries in the predicted depth maps.

The approaches described above all depend on handcrafted features to predict depth values. However, these features are designed beforehand to `describe' a given set of chosen characteristics. As a result, they may fail when transferred to a new environment. 

\subsection{Deep Learning-based Methods}
Motivated by the success of CNN in image classification, dense prediction tasks have also been addressed using CNN-based methods \cite{eigen2014depth, eigen2015predicting, liu2015learning, laina2016deeper, cao2017estimating, ma2018sparse, li2018monocular, xu2018structured, hu2019revisiting, chen2019structure, eom2020temporally, ye2020dpnet, zhou2020padenet, yang2020omnisupervised, liu2020structured}. Eigen et al. \cite{eigen2014depth} introduced the first deep learning-based MDE method. Two deep networks were employed to fulfill the task, the first network predicts a global depth map from the entire image while the second network refines the global prediction at a local level. Eigen and Fergus \cite{eigen2015predicting} extended \cite{eigen2014depth} to multi-task learning. Liu et al. \cite{liu2015learning} developed a deep convolutional neural field (DCNF) network. The designed DCNF combines the strength of deep CNN and continuous CRF in a unified framework. Both \cite{eigen2014depth} and \cite{liu2015learning} use fully connected (FC) layers, which involve a huge number of parameters resulting in expensive computation. \par
To avoid the latency from FC layers, Laina et al. \cite{laina2016deeper} proposed a fully convolutional residual network (FCRN) for MDE. FCRN consists of an encoder and a decoder. The encoder is built on the same basis as the fully convolutional part of ResNet-50, which produces feature maps at 1/32 scale of the input image. The decoder incorporates these feature maps and outputs the final depth map. Furthermore, Laina et al. demonstrated that the depth of the encoder network has great influence on the accuracy of depth estimation, because a deeper network has a larger receptive field. Inspired by this finding, \cite{cao2017estimating, li2018monocular, hu2019revisiting, chen2019structure, ye2020dpnet} utilized CNNs have more than 100 layers in the encoder and produced much better results than their variant with ResNet-50 \cite{he2016deep}. Besides, Hu et al. \cite{hu2019revisiting} and Chen et al. \cite{chen2019structure} also exploited multi-scale features and added a refinement module to their network. \par
Cao et al. \cite{cao2017estimating} formulated depth estimation as a pixelwise classification task. In addition, fully connected CRFs are used as a post-processing operation to further improve the performance. Li et al. \cite{li2018monocular} used side-outputs from the different layers of encoder network to exploit multi-scale features for depth estimation. The above mentioned methods either build on top of the extremely deep CNNs \cite{laina2016deeper, li2018monocular, hu2019revisiting, chen2019structure} or use post-processing operations \cite{cao2017estimating}. Thus, they suffer a heavy computational cost and cannot be run in real-time without high-end GPUs. 

To accelerate the running speed of MDE, Wofk et al. \cite{wofk2019fastdepth} designed an encoder-decoder network by using a light-weight CNN \cite{howard2017mobilenets} as encoder. Moreover, they utilize network pruning techniques \cite{yang2018netadapt} to further reduce the model size. Although the obtained network has achieved great improvement in speed, the accuracy of depth estimation decreased significantly. Recently, Zhou et al. \cite{zhou2020padenet} designed a PADENet for panoramic monocular depth estimation. They trained the PADENet in unsupervised manner in order to overcome the lack of omnidirectional image dataset with groundtruth depth. \par
Motivated by the above mentioned approaches, in this study, we aim to develop a CNN which has a proper trade-off between the accuracy and speed of depth estimation. Specifically, we design the network by assembling two relatively shallow encoder-decoder style subnetworks in a unified framework but do not use any post-processing module. In addition, we build our network on top of a series of simple convolutional layers rather than Inception modules \cite{szegedy2015going} or Residual blocks \cite{he2016deep}. Benefiting from this design, our network has a simple and shallow architecture. Since the unsupervised learning methods normally suffer from lower accuracy, we trained the proposed network on datasets \cite{silberman2012indoor, saxena2008make3d, geiger2013vision, mancini2018j} with the RGB images and ground-truth depth in a supervised fashion in order to acquire better accuracy. \par 
\begin{figure*}[h]
	\centering	   
	\includegraphics[width=.8\linewidth]{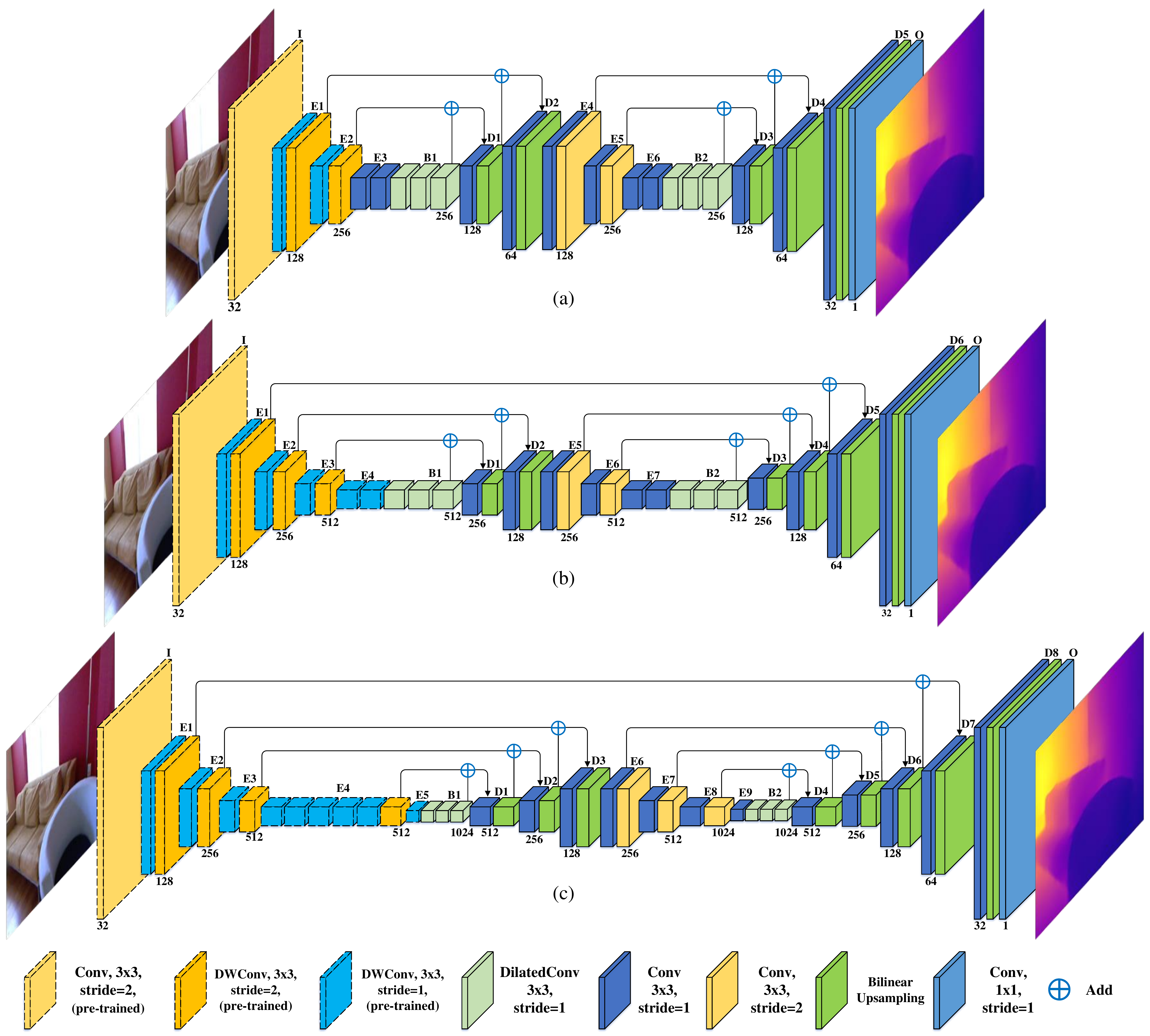}
	\caption{Architecture of the proposed monocular depth estimation networks (best viewed in color). (a) MobileXNet-Small, (b) MobileXNet, and (c) MobileXNet-Large. $I$ represents the input layer, $E_{i}$ means the $i_{th}$ encoder block, $D_{i}$ indicates the $i_{th}$ decoder block and $O$ denotes the output layer.}
	\label{fig:mobilexnet architectures}
\end{figure*}
\section{Methodology}
In this section, we first introduce the details of the proposed CNN architecture for monocular depth estimation (MDE) and then describe the loss functions used for training in this study. 
\subsection{CNN Architecture} 
Targeting at addressing the problems with state-of-the-art approaches and achieving the trade-off between accuracy and speed, we design a novel network which learns the end-to-end mapping from an RGB image to the corresponding depth map. We fulfil the task by assembling two simple and shallow encoder-decoder style subnetworks in a unified framework. As shown in Fig. \ref{fig:mobilexnet architectures}b, each subnetwork only consist of simple convolutional layers. The encoder of the first subnetwork takes a single RGB image as input. After four times downsampling operation, it produces feature maps ($F_{1}$) at $1/16$ scale of the input image. This avoids a greater number of successive downsamplings which impairs the accuracy of recovering boundary level detail and the depth estimation on the small sized images. Subsequently, $F_{1}$ are passed through two upsampling steps and the generated feature maps $F_{2}$ are fed to the second subnetwork, which consists of two downsampling and four upsampling steps and outputs a dense depth map. In order to capture fine-grained image details, skip connections are applied to different levels of the designed network. Instead of concatenating, feature maps are fused via addition to avoid the increase of feature map channels processed in the decoder. This results in a further improvement in running speed. \par
To reduce network latency, we use depthwise separable convolution to design the encoder of the first subnetwork (see Fig. 2b). The depthwise separable convolution factorizes a regular convolution into a depthwise convolution and a pointwise convolution. Specifically, the depthwise convolution applies a single filter at each input channel, and the pointwise convolution is used to create a linear combination of the output of the depthwise layer. \par
The depthwise convolution with one filter per input channel is defined as:
\begin{equation}
{\hat{G}_{k,l,m} = {\sum_{i,j}\hat{K}_{i,j,m}}\cdot{F_{k+i-1,l+j+1,m}}}{,}
\end{equation}
where $\hat{K}$ is the depthwise convolutional kernel with size of ${D_K}\times{D_K}\times{M}$, and the $m_{th}$ filter is operated on the $m_{th}$ channel in $F$ to produce the $m_{th}$ channel of the output feature map $\hat{G}$. Hence, it has a computational complexity of ${{D_{K}}\cdot{D_{K}}\cdot{M}\cdot{D_{F}}\cdot{D_{F}}}$. Since the depthwise convolution only operates on input channels, and is followed by a $1\times1$ convolution which combines the output to generate the final feature map, the computational cost of depthwise separable convolutions is:
\begin{equation}
{{D_{K}}\cdot{D_{K}}\cdot{M}\cdot{D_{F}}\cdot{D_{F}}+{M}\cdot{N}\cdot{D_{F}}\cdot{D_{F}}}{.}
\end{equation}
While for the regular convolution which has the same filter kernel size, the computation cost is  ${{D_{K}}\cdot{D_{K}}\cdot{M}\cdot{N}\cdot{D_{F}}\cdot{D_{F}}}$. By using depthwise separable convolution we can reduce the number of parameters in convolutional layer to:
\begin{equation}
{\frac{{D_{K}}\cdot{D_{K}}\cdot{M}\cdot{D_{F}}\cdot{D_{F}}+{M}\cdot{N}\cdot{D_{F}}\cdot{D_{F}}}{{D_{K}}\cdot{D_{K}}\cdot{M}\cdot{N}\cdot{D_{F}}\cdot{D_{F}}} = \frac{1}{N} + \frac{1}{D_{k}^2}}{.}
\end{equation}
For depthwise separable convolutions with kernel size of 3, the amount of computation is 8 to 9 times less than the regular convolution. Thus, it is helpful to improve the computation speed of MobileXNet. In order to take advantage of the pre-trained weights on the ImageNet dataset \cite{russakovsky2015imagenet}, we design the encoder (boxes with dotted line in Fig. 2b) of the first subnetwork with the same configuration as the first nine layers of MobileNet \cite{howard2017mobilenets}. \par
\begin{figure}[h]
	\centering
	\subfloat[]{
		\includegraphics[height=.25\linewidth]{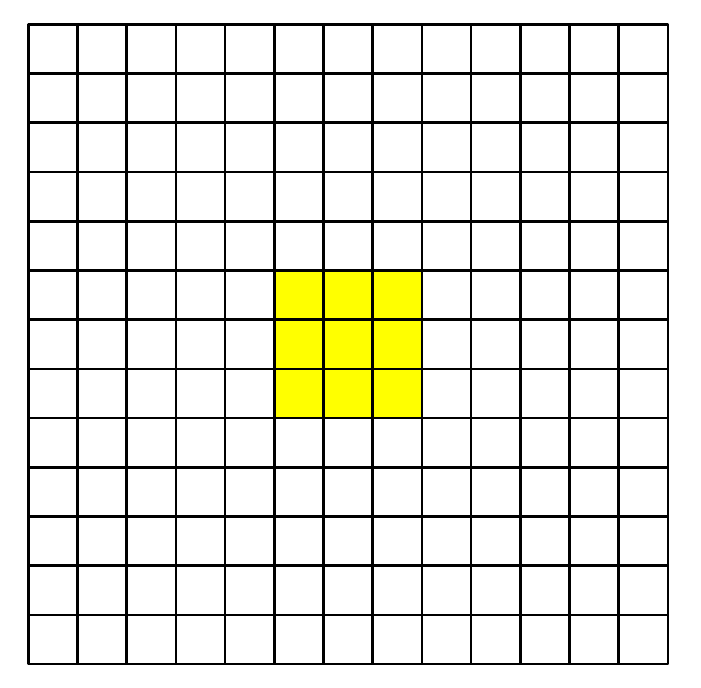}
		\label{fig:1-dilated}}
	\hspace{0.16in}
	\subfloat[]{
		\includegraphics[height=.25\linewidth]{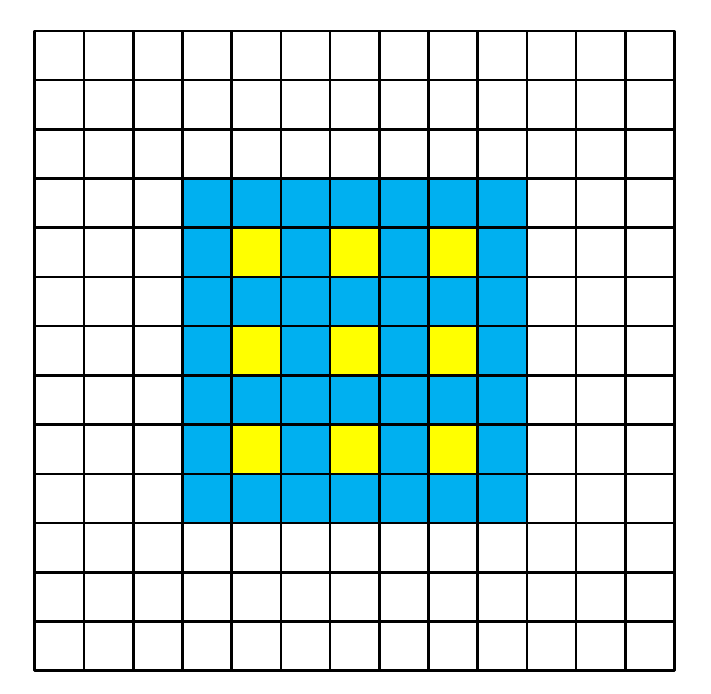}
		\label{fig:2-dilated}}
	\hspace{0.16in}
	\subfloat[]{
		\includegraphics[height=.25\linewidth]{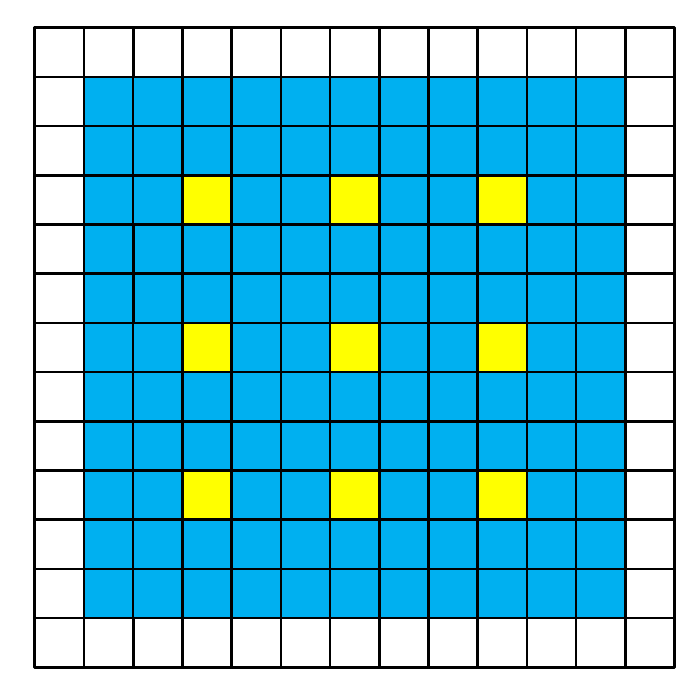}
		\label{fig:3-dilated}}
	\vspace{0.051in}
	\caption{Illustration of the dilated convolution. (a), (b), (c) are 1-dilated, 2-dilated, 3-dilated convolution respectively.}
	\label{F:dilated}
	\vspace{-0.1in}
\end{figure}
A larger receptive field enables the network to capture more context information, which can be leveraged to improve depth estimation performance \cite{laina2016deeper}. The dilated convolution expands the receptive field of the neuron without increasing the number of model parameters \cite{yu2015multi}. Therefore, we insert a bridge module between the encoder and decoder in each subnetwork. Specifically, the bridge module consists of three dilated convolutional layers. Given a discrete function defined by ${\mathcal{F}}:{\mathcal{Z}^2}{\rightarrow}{\mathcal{R}}$, let ${{\Omega}_{r} = {[-r, r]^2} \cap {\mathcal{Z}^2}}$ and ${k}: {{\Omega}_{r}}{\rightarrow}{\mathcal{R}}$ be a discrete filter of size $(2r+1)^2$. The discrete convolution operator $*$ can be expressed as:
\begin{equation}
{(F * k)(p) = {\sum_{s+t=p}F(s)k(t)}}{.}
\end{equation}
Based on this filter, the dilated convolution with a dilation factor $r$ can be formulated as:
\begin{equation}
{(F *_{r} k)(p) = {\sum_{s+{lt}=p}F(s)k(t)}}{,}
\end{equation}
where $*_{r}$ is an $r$-dilated convolution. The regular convolution $*$ is simply the 1-dilated convolution. The illustration of dilated convolution is shown in Fig. \ref{F:dilated}. In this study, the dilation rates are set as 1, 2 and 3. By doing this, the decoders can get the context information captured in multiple scales. \par
The task of the decoder is to refine and upsample the output of the encoder. The core part of the decoder is the upsample layer, such as unpooling, up-projection, transpose convolution and interpolation combined with convolution. According to \cite{laina2016deeper}, up-projection produced better performance than unpooling and transpose convolution. However due to the relatively complex architecture, it significantly increases the number of network parameters. Inspired by \cite{wojna2019devil}, we choose the combination of a bilinear interpolation and a $3\times3$ convolutional layer as the upsample layer in this study. The $3\times3$ convolutional layer reduces the number of output channels to half of the number of input channels. The bilinear interpolation increases the spatial resolution of intermediate feature maps to the same size as the output of the corresponding encoder layer. The main advantage of this design is the simple network architecture and reduced computational time. \par
We also design two variants that perform three and five times of downsampling in the first subnetwork, and we name them as MobileXNet-Small and MobileXNet-Large respectively. We train and test MobileXNet and the variants on the NYU depth v2 dataset \cite{silberman2012indoor} to compare the performance. The network architectures are shown in Fig. \ref{fig:mobilexnet architectures}.
\subsection{Loss Functions}
A loss function is applied to measure how far the predicted depth map is from the groundtruth, and this is used to supervise the update of network weights. Thus, it plays an important role in training CNNs. A commonly used loss function for depth estimation is the $L_{2}$ loss as it is more effective for pixels with larger error \cite{irie2019robust}. The $L_{2}$ loss is defined as:
\begin{equation}
    {{L}_{2}(d, d^{*}) = \frac{1}{N} \sum_{i}^{N} |d_{i} - d_{i}^{*}|^2}{,}
\end{equation}
where $N$ is the number of pixels being considered, $d$ and $d^{*}$ are the predicted and groundtruth depth respectively. In \cite{laina2016deeper}, Laina et al. introduce a reverse Huber loss: 
\begin{equation}
    {L}_{berHu}(d, d^{*}) = \begin{cases}
    |d-d^{*}| & if |d-d^{*}| {\le} c{,} \\
    {\frac{(d-d^{*})^2 + c^{2}}{2c}} & otherwise{,}
                            \end{cases}
\end{equation}
where $c$ is the threshold and set as $c = \frac{1}{5}{max_{i}(|d-d^{*}|)}$. The berHu loss is equal to $L_{2}$ loss when the error exceeds $c$, whilst when the error falls in $[-c, c]$ the berHu loss is equal to the mean absolute error ($L_{1}$ loss): 
\begin{equation}
{{L}_{1}(d, d^{*}) = \frac{1}{N} \sum_{i}^{N} |d_{i} - d_{i}^{*}|}{,}
\end{equation}
which is more robust to outliers in the dataset \cite{kumari2019autodepth}. However, the above-mentioned loss functions are sensitive to errors emerging at step edges, such as difference between sharp and blurry edges \cite{jiao2018look}. To penalize the errors around edges, we define a hybrid loss which combines the regular $L_{1}$ loss with the image gradient-based $L_{1}$ loss:
\begin{equation}
{L_{grad}(d, d^{*}) = \frac{1}{N} \sum_{i}^{N} |{\bigtriangledown}_{x}(d_{i}, d_{i}^{*})| + |{\bigtriangledown}_{y}(d_{i}, d_{i}^{*})|}{,}
\end{equation}
where ${\bigtriangledown}_{x}$ and ${\bigtriangledown}_{y}$ are spatial derivative in $x$ and $y$ direction respectively. Finally, the defined hybrid loss is formulated as:
\begin{equation}
{L_{hybrid} = L_{depth} + L_{grad}}{,}
\end{equation}
where $L_{depth}$ is the regular $L_{1}$ loss, and $L_{grad}$ is the image gradient-based $L_{1}$ loss.
\section{Experimental Setup}
We introduce our experimental setup in this section. It includes three parts: implementation details, data augmentation methods and performance metrics.
\subsection{Implementation Details}
We implement the proposed networks using PyTorch. A desktop with an i7-7700 CPU, 16GB RAM and a single Nvidia GTX 1080 GPU is used for training and testing. The weights of the encoder of the first subnetwork of MobileXNet and its variants are initialized with the pre-trained weights. The other layers are initialized from the Gaussian distribution with a standard deviation of $\sqrt{2/N}$, where $N$ represents the number of incoming nodes of one neuron. The training is optimized by using the SGD optimizer with an initial learning rate of 0.01, and the batch size is set as 8. For the NYU depth v2 \cite{silberman2012indoor}, KITTI \cite{geiger2013vision} and Unreal \cite{mancini2018j} dataset, we train the network for 20 epochs, and reduce the learning rate to $20\%$ every 5 epochs. For the Make3D dataset \cite{saxena2008make3d}, we train MobileXNet for 100 epochs, and reduce the learning rate to $20\%$ every 40 epochs. \par
\subsection{Data Augmentation}
To increase the diversity of training samples, we utilize the following data augmentation methods, which are applied to each RGB and depth image pair in an online fashion:

(1) Random rotation: RGB and depth image pairs are rotated by a random angle $r\in[-5, 5]$ degrees.

(2) Random scale: RGB images are scaled by a random factor $s\in[1, 1.5]$, and the corresponding depth maps are divided by $s$. 

(3) Color Jitter: the brightness, contrast, and saturation of the RGB images are scaled by a random factor $c\in[0.6, 1.4]$.

(4) Random flips: RGB and depth image pairs are horizontally flipped with the probability of 0.5.
\subsection{Performance Metrics}
In this work, we evaluate each method with the following metrics:
 
(1) The Root Mean Square Error (RMSE): 
\begin{equation}
    {RMSE = \sqrt{\frac{1}{N}\sum_{i}^{N}|d_{i} - d_{i}^{*}|^2}}{.}
\end{equation}

(2) Mean Relative Error (REL): 
\begin{equation}
    {REL = \frac{1}{N}\sum_{i}^{N}\frac{|d_{i} - {d_{i}^{*}}|}{{d_{i}}}}{.}
\end{equation}

(3) Thresholded Accuracy ($\delta$): it is the percentage of predicted pixels where the relative error is within a threshold. The formula is represented as follows:
\begin{equation}
    {max(\frac{d^{*}}{d}, \frac{d}{d^{*}}) < \delta_{i}, \delta_{i} = 1.25^{i}}{.}
\end{equation} 

(4) Running time ($t_{GPU}$): the average execution time of testing each frame on a single GPU, we report the running time of our method on the GTX 1080 GPU. Due to the code and GPU model of some methods not being available to us, we cannot do a direct comparison on the Nvidia GPU. Therefore, we instead compare against their running time and GPU model as reported in the literature.
\section{Experimental Results}
In this section, we present the experimental results on monocular depth estimation (MDE). We evaluate the proposed method on both indoor and outdoor scenes. Four data sets, the NYU depth v2 \cite{silberman2012indoor}, KITTI \cite{geiger2013vision}, Make3D \cite{saxena2008make3d} and Unreal \cite{mancini2018j} are selected as benchmarks. Specifically, the NYU depth v2 dataset was captured in indoor environments, the KITTI \cite{geiger2013vision} and Make3D dataset \cite{saxena2008make3d} were collected in real-world outdoor scenes, while the UnrealDataset \cite{mancini2018j} was gathered from simulated outdoor surroundings. We first evaluate MobileXNet with different loss functions and network configurations on the NYU depth v2 dataset \cite{silberman2012indoor}. Then, we compare it with state-of-the-art methods on four benchmarks. \par
\begin{table*}[h]
	\renewcommand{\arraystretch}{1.3}	
	\caption{Evaluation of loss functions (Rows 1-4) and dilation rates (Rows 4-8) on the NYU depth v2 dataset \cite{silberman2012indoor}. {$[1, 2, 3]^{*}$} means depth-wise separable convolutions with dilations. The \textcolor{red}{red} and \textcolor{red}{\textbf{bold}} values indicate the best results.}
	\centering
	\label{table: loss_function}	
	\begin{tabular}{ c | c | c | c | c | c | c | c | c | c } 
		\hline
		Row & Method & Loss & Dilation Rates & \cellcolor{gray} \makecell{RMSE} & \cellcolor{gray} \makecell{REL} & {\cellcolor{pink} \makecell{${{\delta}_1}$}} & \cellcolor{pink}{\makecell{${{\delta}_2}$}} & {\cellcolor{pink} \makecell{${{\delta}_3}$}} & \cellcolor{gray} \makecell{$t_{GPU}$} \\ 
		\hline\hline
		
		1 & MobileXNet & ${\mathcal{L}_{1}}$ & [1, 2, 3] & 0.558 & 0.158 & 0.784 & 0.945 & 0.984 & 8.1 ms \\
		2 & MobileXNet & ${\mathcal{L}_{2}}$ & [1, 2, 3] & 0.553 & 0.158 & 0.777 & 0.948 & 0.986 & 7.8 ms \\     
		3 & MobileXNet & \makecell{berHu} & [1, 2, 3] & 0.550 & 0.160 & 0.779 & 0.947 & 0.987 & 8.9 ms \\   
		4 & MobileXNet & \makecell{Hybrid} & [1, 2, 3] & 0.537 & \textcolor{red}{\textbf{0.146}} & \textcolor{red}{\textbf{0.799}} & \textcolor{red}{\textbf{0.951}} & \textcolor{red}{\textbf{0.988}} & 7.9 ms \\	 
		5 & MobileXNet & \makecell{Hybrid} & [1, 1, 1] & 0.552 & 0.156 & 0.785 & 0.946 & 0.985 & 6.8 ms \\
		6 & MobileXNet & \makecell{Hybrid} & [2, 2, 2] & 0.545 & 0.156 & 0.794 & 0.947 & 0.987 & 9.0 ms \\
		7 & MobileXNet & Hybrid & [3, 3, 3] & \textcolor{red}{\textbf{0.533}} & 0.148 & 0.797 & \textcolor{red}{\textbf{0.951}} & 0.987 & 8.9 ms \\
		8 & MobileXNet & Hybrid & {$[1, 2, 3]^{*}$} & 0.552 & 0.162 & 0.786 & 0.942 & 0.984 & \textcolor{red}{\textbf{6.2 ms}} \\			 		 	
		\hline  
	\end{tabular}
	
	\begin{tabular}{| c | c | c |} 
		{\cellcolor{gray} Lower is better} & {\cellcolor{pink} Higher is better} & {$\delta_1: \delta < 1.25$}, {$\delta_2: \delta < 1.25^{2}$}, {$\delta_3: \delta < 1.25^{3}$} \\ \hline 
	\end{tabular} 
	
\end{table*}
\begin{table*}[h]
	\renewcommand{\arraystretch}{1.3}	
	\caption{Evaluation of the weight initialization and convolution types in the encoder of the first sub-network (Rows 1-3) and the second sub-network (Rows 4-6) on the NYU depth v2 dataset \cite{silberman2012indoor}. DwConv indicates the depthwise separable convolutional layers initialized from the pre-trained weights, DwConv-IG represents the depthwise separable convolutional layers initialized from the Gaussian distribution, and Conv-IG means the regular convolutional layers initialized from the Gaussian distribution. The \textcolor{red}{red} and \textcolor{red}{\textbf{bold}} values indicate the best results.}
	\centering
	\label{table: weight_initialization}	
	\begin{tabular}{ c | c | c | c | c | c | c | c | c } 
		\hline
		Row & Method & {Convolution Types} & \cellcolor{gray} \makecell{RMSE} & \cellcolor{gray} \makecell{REL} & {\cellcolor{pink} \makecell{${{\delta}_1}$}} & \cellcolor{pink}{\makecell{${{\delta}_2}$}} & {\cellcolor{pink} \makecell{${{\delta}_3}$}} & \cellcolor{gray} \makecell{$t_{GPU}$} \\ 
		\hline\hline		
		1 & MobileXNet & DwConv & \textcolor{red}{\textbf{0.537}} & \textcolor{red}{\textbf{0.146}} & \textcolor{red}{\textbf{0.799}} & \textcolor{red}{\textbf{0.951}} & \textcolor{red}{\textbf{0.988}} & 7.9 ms \\	 
		2 & MobileXNet & DwConv-IG & 0.669 & 0.198 & 0.692 & 0.909 & 0.973 & 8.5 ms \\
		3 & MobileXNet & Conv-IG & 0.632 & 0.186 & 0.724 & 0.923 & 0.976 & 9.3 ms \\
		\hline \hline
		4 & MobileXNet & DwConv & 0.822 & 0.209 & 0.584 & 0.862 & 0.961 & \textcolor{red}{\textbf{6.8 ms}} \\		
		5 & MobileXNet & DwConv-IG & 0.550 & 0.156 & 0.788 & 0.947 & 0.986 & 7.3 ms \\
		6 & MobileXNet & Conv-IG & \textcolor{red}{\textbf{0.537}} & \textcolor{red}{\textbf{0.146}} & \textcolor{red}{\textbf{0.799}} & \textcolor{red}{\textbf{0.951}} & \textcolor{red}{\textbf{0.988}} & 7.9 ms \\							 		 	
		\hline  
	\end{tabular}
	
	\begin{tabular}{| c | c | c |} 
		{\cellcolor{gray} Lower is better} & {\cellcolor{pink} Higher is better} & {$\delta_1: \delta < 1.25$}, {$\delta_2: \delta < 1.25^{2}$}, {$\delta_3: \delta < 1.25^{3}$} \\ \hline 
	\end{tabular} 
	
\end{table*}
\subsection{NYU Depth Dataset}
The NYU depth v2 dataset \cite{silberman2012indoor} consists of about 240k RGB and depth image pairs captured from 464 different indoor scenes through a Microsoft Kinect camera. In this study, we train the designed method on about 48k synchronized RGB and depth images pairs, and test it on 654 images. Both training and testing data are released by \cite{wofk2019fastdepth}. Following \cite{eigen2014depth, laina2016deeper, wofk2019fastdepth}, we first downsample the original images from $640\times480$ pixels to half size, and then center crop $304\times228$ pixels region as input to the network. \par
\subsubsection{Evaluation of Loss Functions} In this subsection, we perform a set of experiments over different loss functions based on the designed MobileXNet. Four loss functions described in Section III-B are tested, and the results are listed in the first 4 rows of Table \ref{table: loss_function}. It can be observed that both the berHu loss and the $L_{2}$ loss outperform the $L_{1}$ loss. In addition, the berHu loss performs slightly better than the $L_{2}$ loss. It should be noted that the designed hybrid loss which combines the regular $L_{1}$ loss with the image gradient-based $L_{1}$ loss generates the best performance. Therefore, we use the \textsl{hybrid} loss function as the default loss in this study.
\subsubsection{Evaluation of Dilation Rate}
In this subsection, we evaluate the performance of different dilation rate configurations in the bridge modules. The dilation rates in the bridge modules are set as [1, 1, 1], [2, 2, 2], [3, 3, 3] and [1, 2, 3], while [1, 1, 1] means the bridge modules have three regular convolutional layers. The results produced from these 4 configurations are listed in Rows 4 to 7 of Table \ref{table: loss_function}. \par 
As can be seen, (1) the dilated convolutions improve the performance in both accuracy and error metrics. We attribute this to the fact that it has a larger receptive field and captures more context information. It should be noted that we do not further increase the dilation rate, as we did not obtain empirical improvement when dilation rate was larger than 3; (2) the configuration of [3, 3, 3] produces the best RMSE value, while the values of REL, $\delta_1$, $\delta_3$ and running time are inferior to [1, 2, 3]. It demonstrates that using multiple dilated convolutional layers with different dilation rates to capture the multiple-scale context information is helpful in improving accuracy. Besides, the MobileXNet using the bridge modules with dilation rates of [1, 2, 3] runs faster than [3, 3, 3] by 1ms. This suggests that [1, 2, 3] achieves better accuracy and speed balance; (3) we further apply depthwise separable convolutions with dilation rates of 1, 2 and 3 to the bridge modules. According to the 4th and 8th rows of Table \ref{table: loss_function}, the depthwise separable convolution lowers the running time by 1.7ms but the performance also decreased. However, the MobileXNet using the regular convolutions with dilation rates of [1, 2, 3] generates the best performance and adequate speed for real-time application. Thus, we choose it as the default configuration. \par
\begin{table*}[h]
	\renewcommand{\arraystretch}{1.3}
	\caption{Comparison of the proposed MobileXNet against different variants and U-Net \cite{ronneberger2015u} on the NYU depth v2 dataset \cite{silberman2012indoor}. {$\triangle$} represents the first 9 layers of MobileNet, {$\Box$} denotes the first 7 layers of MobileNet, {$\diamondsuit$} means the network does not use CNN designed for image classification. M and G indicate $\times 10^{6}$ and $\times 10^{9}$ respectively. The \textcolor{red}{red} and \textcolor{red}{\textbf{bold}} values indicate the best results.}
	\label{table: architecture}
	\centering
	\begin{tabular}{c | c | c | c | c | c | c | c | c | c | c | c } 
		\hline
		Row & Method & Backbone & Parameters & Flops & Memory & \cellcolor{gray} \makecell{RMSE} & \cellcolor{gray} \makecell{REL} & {\cellcolor{pink} \makecell{${{\delta}_1}$}} & {\cellcolor{pink} \makecell{${{\delta}_2}$}} & {\cellcolor{pink} \makecell{${{\delta}_3}$}} & {\cellcolor{gray} \makecell{$t_{GPU}$}} \\ 
		\hline\hline
		1 & MobileXNet & $\triangle$ & 24.95 M & 9.78 G & 111.12 MB & \textcolor{red}{\textbf{0.537}} & \textcolor{red}{\textbf{0.146}} & \textcolor{red}{\textbf{0.799}} & 0.951 & \textcolor{red}{\textbf{0.988}} & 7.9 ms \\	 
		2 & MobileXNet-Small & $\Box$ & 6.51 M & 8.34 G & 104.61 MB & 0.606 & 0.180 & 0.733 & 0.930 & 0.981 & \textcolor{red}{\textbf{6.1 ms}} \\    
		3 & MobileXNet-Large & \makecell{MobileNet} & 91.07 M & 11.85 G & 121.17 MB & 0.538 & 0.148 & \textcolor{red}{\textbf{0.799}} & \textcolor{red}{\textbf{0.952}} & 0.987 & 15.2 ms \\
		4 & ShuffleXNet & ShuffleNetV2 & 5.20 M & 1.73 G & \textcolor{red}{\textbf{41.93 MB}} & 0.580 & 0.163 & 0.766 & 0.943 & 0.985 & 6.9 ms \\
		5 & EfficientXNet & EfficientNet-B0 & \textcolor{red}{\textbf{1.96 M}} & \textcolor{red}{\textbf{0.39 G}} & 74.19 MB & 0.575 & 0.162 & 0.766 & 0.946 & 0.985 & 8.2 ms \\		 
		6 & U-Net \cite{ronneberger2015u} (Bilinear) & $\diamondsuit$ & 17.27 M & 42.22 G & 401.05 MB & 0.705 & 0.206 & 0.661 & 0.900 & 0.972 & 14.1 ms \\ 
		7 & U-Net \cite{ronneberger2015u} (DeConv) & $\diamondsuit$ & 31.04 M & 48.55 G & 426.32 MB & 0.726 & 0.212 & 0.644 & 0.892 & 0.971 & 19.8 ms \\ 
		\hline 
	\end{tabular}
	
	\begin{tabular}{| c | c | c |} 
		\cellcolor{gray}Lower is better & \cellcolor{pink}Higher is better & {$\delta_1: \delta < 1.25$}, {$\delta_2: \delta < 1.25^{2}$}, {$\delta_3: \delta < 1.25^{3}$} \\ \hline 
	\end{tabular} 
	
\end{table*}
\subsubsection{Effect of Weight Initialization} To evaluate the effect of weight initialization in the backbone (the \textbf{encoder} of the \textbf{first} subnetwork) of MobileXNet, we initialized it with the pre-trained MobileNet model (row 1, Table \ref{table: weight_initialization}) and Gaussian distribution (row 2, Table \ref{table: weight_initialization}) respectively. Furthermore, we design a variant of MobileXNet by replacing the depthwise separable convolutional layers in the backbone with regular convolutional layers (row 3, Table \ref{table: weight_initialization}). Since the ImageNet dataset pre-trained weights of the regular convolutional layers are not available, we initialize them from the Gaussian distribution. As can be observed from the \textsl{2nd} and \textsl{3rd} rows of Table \ref{table: weight_initialization}, when initialized from the Gaussian distribution, the variant with regular convolutional layers (row 3, Table \ref{table: weight_initialization}) outperforms the Gaussian distribution initialized MobileXNet (row 2, Table \ref{table: weight_initialization}) in RMSE, REL, $\delta_1$, $\delta_2$ and $\delta_3$, while the running speed is inferior. We attribute this to the fact that the regular convolution has more parameters than the depthwise separable convolution. However, when initializing the backbone from the pre-trained weights (row 1, Table \ref{table: weight_initialization}), the performance of MobileXNet improved significantly. Thus, we choose the depthwise separable convolution in the backbone and initialize it from the pre-trained weights in this study. \par
\begin{figure}[h]
	\centering
	\includegraphics[width=.9\linewidth]{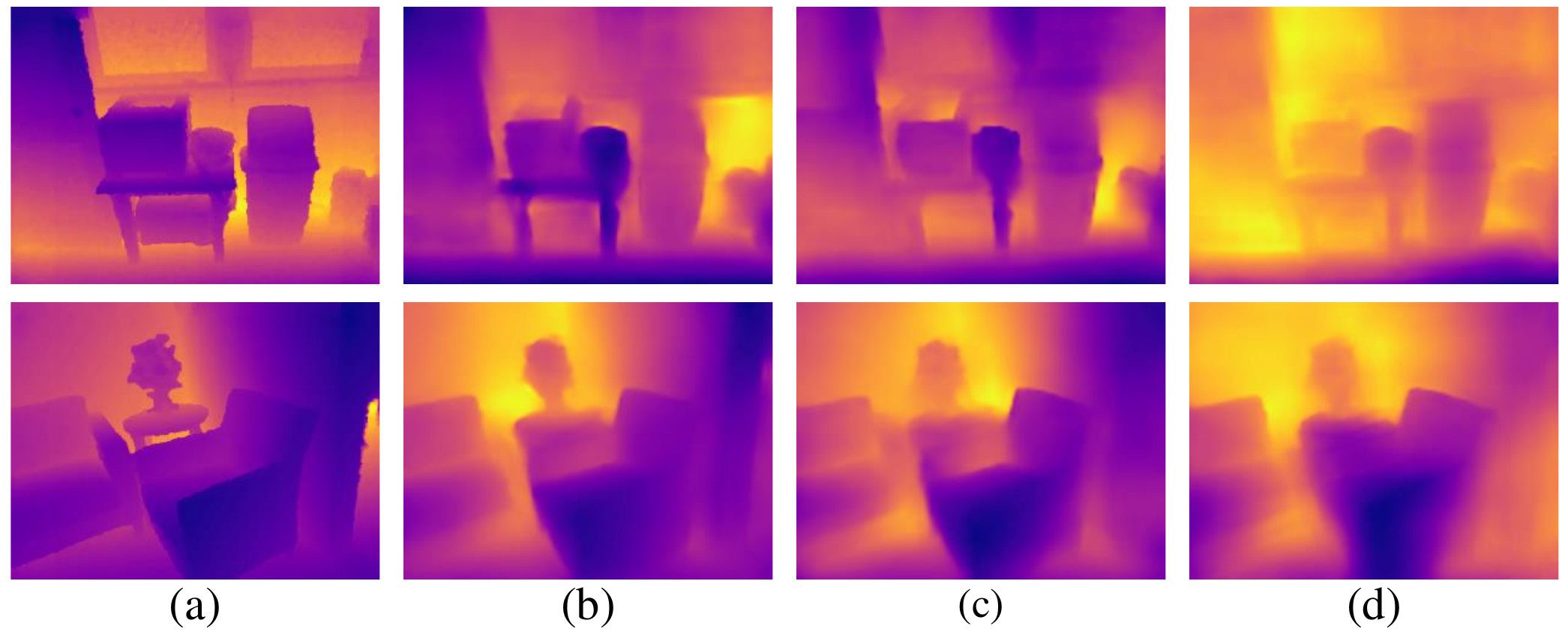} \\
	\caption{Qualitative results of MobileXNet with different weight initialization and convolution type in the encoder of the second sub-network. (a) Groundtruth, (b) Conv-IG, (c) DWConv-IG, and (d) DWConv. Color represents depth (yellow is far, blue is close).}
	\label{fig:failure_mode}
\end{figure}
With the backbone initialized from the pre-trained MobileNet model, we further design the \textbf{encoder} of the \textbf{second} subnetwork (\textsl{E5-E7} blocks in Fig. \ref{fig:mobilexnet architectures}b) with depthwise separable convolutional layers and initialize it with the pre-trained weights of the corresponding layers of MobileNet \cite{howard2017mobilenets}. Experimental results are shown in rows 4 to 6 of Table \ref{table: weight_initialization}. Figure \ref{fig:failure_mode} also displays the qualitative results. \par

It can be seen from Table \ref{table: weight_initialization} that when we initialized the encoder of the second subnetwork with the pre-trained weights (row 4, Table \ref{table: weight_initialization}), MobileXNet does not provide good performance. Furthermore, the produced depth maps are blurry and suffer from sharp discontinuities of shapes of objects (see Fig. \ref{fig:failure_mode}d). We attribute the bad performance to the following facts. In essence, the MobileXNet stacks two encoder-decoder style sub-networks. The encoder and decoder of the first sub-network are trained to describe input images and generate feature maps with $1/4$ size of the input image, respectively. The produced feature maps are passed to the second subnetwork in order to produce depth maps. The second subnetwork works as a learned refining module which operates with downsampling and upsampling steps. This subnetwork tends to be more dependent on the training dataset than the backbone which is utilized to learn generic features \cite{dong2020defect}. The ImageNet dataset \cite{russakovsky2015imagenet} used for pre-training the MobileNet is different from the NYU depth v2 dataset. Hence, initializing the encoder of the second subnetwork impairs the performance of depth estimation. \par
To further augment our experiments, we then initialize the parameters of the depthwise separable convolutional layers in the encoder of the second subnetwork with a Gaussian distribution (row 5, Table \ref{table: weight_initialization}). As shown in Table \ref{table: weight_initialization}, initializing the encoder of the second subnetwork from the Gaussian distribution outperforms initialization from pre-trained weights. Compared with the original MobileXNet (row 6, Table \ref{table: weight_initialization}), the MobileXNet with the depthwise separable convolution in the encoder of the second subnetwork has inferior performance. In addition, the boundary of small or thin objects in the produced depth maps (Fig. \ref{fig:failure_mode}c) are unclear. Therefore, we use regular convolutional layers in the encoder of the second subnetwork and initialize it from the Gaussian distribution. \par
\begin{table*}[h]
	\renewcommand{\arraystretch}{1.3}
	\caption{Comparison of performances on the NYU depth v2 dataset \cite{silberman2012indoor}. {$\diamondsuit$} means the network does not use CNN designed for image classification, {$\triangle$} represents the first 9 layers of MobileNet. The \textcolor{red}{red} and \textcolor{red}{\textbf{bold}} values indicate the best results.}
	\centering
	\label{table: comparison_on_nyu}
	\begin{tabular}{ c | c | c | c | c | c | c | c | c | c } 
		\hline
		Row & Method & Backbone & \cellcolor{gray} \makecell{RMSE} & \cellcolor{gray} \makecell{REL} & {\cellcolor{pink} \makecell{${{\delta}_1}$}} & {\cellcolor{pink} \makecell{${{\delta}_2}$}} & {\cellcolor{pink} \makecell{${{\delta}_3}$}} & \cellcolor{gray} \makecell{$t_{GPU}$} & Device \\
		\hline\hline
		1 & Eigen et al. \cite{eigen2014depth} & $\diamondsuit$ & 0.907 & 0.215 & 0.611 & 0.887 & 0.971 & N/A & N/A \\ 
		2 & Eigen and Fergus \cite{eigen2015predicting} & VGG-16 & 0.641 & 0.158 & 0.769 & 0.950 & 0.988 & N/A & N/A \\
		3 & Liu et al. \cite{liu2015learning} & VGG-16 & 0.759 & 0.213 & 0.650 & 0.906 & 0.976 & N/A & N/A \\
		4 & Laina et al. \cite{laina2016deeper} (UpConv) & ResNet-50 & 0.604 & 0.132 & 0.789 & 0.946 & 0.986 & 78 ms & Nvidia GTX Titan \\  
		5 & Laina et al. \cite{laina2016deeper} (UpProj) & ResNet-50 & 0.573 & 0.127 & 0.811 & 0.953 & 0.988 & 55 ms & Nvidia GTX Titan \\ 		
		6 & Cao et al. \cite{cao2017estimating} & ResNet-152 & 0.540 & 0.141 & \textcolor{red}{\textbf{0.819}} & \textcolor{red}{\textbf{0.965}} & \textcolor{red}{\textbf{0.992}} & N/A & N/A \\  
		7 & Li et al. \cite{li2018monocular} & ResNet-50 & 0.601 & 0.147 & 0.808 & 0.957 & 0.985 & N/A & N/A \\
		8 & He et al. \cite{he2018learning} & VGG-16 & 0.572 & 0.151 & 0.789 & 0.948 & 0.986 & N/A & N/A \\				
		9 & Xu et al. \cite{xu2018structured} & ResNet-50 & 0.593 & \textcolor{red}{\textbf{0.125}} & 0.806 & 0.952 & 0.986 & 150 & Nvidia Titan-X \\
		10 & Wofk et al. \cite{wofk2019fastdepth} (Original) & MobileNet & 0.579 & 0.164 & 0.772 & 0.938 & 0.982 & 5.0 ms & Nvidia GTX 1080 \\ 
		11 & Wofk et al. \cite{wofk2019fastdepth} (Final) & MobileNet & 0.604 & 0.165 & 0.771 & 0.937 & 0.980 & 4.0 ms & Nvidia GTX 1080 \\
		12 & Yang et al. \cite{yang2019fast} & ResNet-18 & 0.628 & 0.199 & 0.708 & 0.916 & 0.975 & 10 ms & Nvidia Titan-X \\
		13 & Hambarde and Murala \cite{hambarde2020s2dnet} & $\diamondsuit$ & 0.543 & 0.160 & 0.773 & 0.959 & 0.989 & N/A & N/A \\		
		14 & Spek et al. \cite{spek2018cream} & ERFNet & 0.687 & 0.190 & 0.704 & 0.917 & 0.977 & \textcolor{red}{\textbf{3.2 ms}} & Nvidia GTX 1080Ti \\		
		15 & MobileXNet (Ours) & $\triangle$ & \textcolor{red}{\textbf{0.537}} & 0.146 & 0.799 & 0.951 & 0.988 & 7.9 ms & Nvidia GTX 1080 \\ \hline
	\end{tabular}
	
	\begin{tabular}{| c | c | c |} \hline
		\cellcolor{gray}Lower is better & \cellcolor{pink}Higher is better & {$\delta_1: \delta < 1.25$}, {$\delta_2: \delta < 1.25^{2}$}, {$\delta_3: \delta < 1.25^{3}$} \\ \hline 
	\end{tabular} 
	
\end{table*}
\subsubsection{Evaluation of Network Architectures}
To validate the effectiveness of the MobileXNet, we compare it with different variants. In addition to the MobileXNet-Small and MobileXNet-Large, we build two variants based on the ShuffleNetV2 \cite{ma2018shufflenet} and EfficientNet \cite{tan2019efficientnet} backbones. The corresponding variants are named as ShuffleXNet\footnote{The conv1 to stage 3 layers of ShuffleNetV2-1{x} \cite{ma2018shufflenet} are used as backbone} and EfficientXNet\footnote{The stage 1 to stage 6 layers of EfficientNet-B0 \cite{tan2019efficientnet} are used as backbone} respectively. The standard encoder-decoder network, U-Net, is used as the baseline. Unlike \cite{ronneberger2015u}, the U-Net includes batch normalization in this work. All methods were trained with the hybrid loss. The experimental results are shown in Table \ref{table: architecture}. \par

It can be observed that: (1) among all methods, the MobileXNet generates the best error and accuracy metric results and its running time is only inferior to the fastest method by 1.8ms; (2) regarding the RMSE, REL, $\delta_1$, $\delta_2$ and $\delta_3$ metrics, the performance of MobileXNet-Large is very close to MobileXNet. However, due to its deep network architecture, the running time is almost doubled; (3) the MobileXNet-Small yields the fastest speed, while its error and accuracy metric results are inferior to its counterparts except the U-Net \cite{ronneberger2015u}; (4) the network parameters, flops and memory footprint of ShuffleXNet are about $4\times$, $5\times$ and $2\times$ fewer than MobileXNet, but it only faster than MobileXNet by 1ms. Moreover, the performance of ShuffleXNet with respect to the error and accuracy metrics is not comparable to MobileXNet; 5) the EfficientXNet is optimal in terms of network parameters and flops, while its running time is longer than MobileXNet. This should be attributed to the fact that the EfficientXNet is built on top of EfficientNet \cite{tan2019efficientnet}, which is optimized for parameter and flops efficiency. Furthermore, EfficientNet scales the feature map resolution and depth at the same time, this leads to the slow GPU inference time \cite{radosavovic2020designing}; and (6) the performance of U-Net \cite{ronneberger2015u} is inferior, no matter which upsampling method is used. In addition, it is the slowest especially when using transpose convolution (DeConv) for upsampling. It should be noted that the weights of U-Net are initialized from a Gaussian distribution. \par 
The MobileXNet generates the best performance in terms of error and accuracy metrics and achieves real-time speed about 126 fps on a GTX 1080 GPU, which is adequate for autonomous driving and robotic applications. In this context, we choose MobileXNet as the network in this study. In the following part we compare its performance with the state-of-the-art. \par 
\begin{table}[h]
	\renewcommand{\arraystretch}{1.3}
	\caption{Comparison of perfrmance with respect to the benefit of data augmentation. The \textcolor{red}{red} and \textcolor{red}{\textbf{bold}} values indicate the best results.}
	\label{table: data_augmentation}
	\centering
		\begin{tabular}{ c | c | c | c | c | c } 
		\hline
		Training Data & \cellcolor{gray} \makecell{RMSE} & \cellcolor{gray} \makecell{REL} & {\cellcolor{pink} \makecell{${{\delta}_1}$}} & {\cellcolor{pink} \makecell{${{\delta}_2}$}} & {\cellcolor{pink} \makecell{${{\delta}_3}$}} \\ 
		\hline\hline
		Original & 0.555 & 0.158 & 0.782 & 0.946 & 0.985 \\ 
		Augmented & \textcolor{red}{\textbf{0.537}} & \textcolor{red}{\textbf{0.146}} & \textcolor{red}{\textbf{0.799}} & \textcolor{red}{\textbf{0.951}} & \textcolor{red}{\textbf{0.988}} \\
		\hline 
		\end{tabular}
	\begin{tabular}{| c | c |} 
	\cellcolor{gray}Lower is better & \cellcolor{pink}Higher is better  \\ \hline 
	\end{tabular} 	
	\end{table}
\subsubsection{Effect of Data Augmentation}
We employ data augmentation to increase the diversity of the training data to enable the trained network to have a better depth estimation performance. We train our MobileXNet on the original training data and the training data with online data augmentation to analyze the benefit of data augmentation. The standard 654 testing images are used as testing data. Experimental results are listed in the last two rows of Table \ref{table: data_augmentation}. It can be observed that data augmentation improves the monocular depth estimation performance, especially the RMSE, REL and $\delta_1$ metrics. Hence, we use data augmentation in this study.
\subsubsection{Comparison with the State-of-the-Art}
In this subsection, we compare MobileXNet with state-of-the-art methods \cite{eigen2014depth, eigen2015predicting, liu2015learning, laina2016deeper, cao2017estimating, li2018monocular, he2018learning, spek2018cream, xu2018structured, wofk2019fastdepth, yang2019fast, hambarde2020s2dnet}. Since we focus on depth estimation from single RGB images, thus, methods that fuse sparse depth points are not compared in this study. The results of \cite{eigen2014depth, eigen2015predicting, liu2015learning, laina2016deeper, cao2017estimating, xu2018structured, li2018monocular, he2018learning, spek2018cream, yang2019fast, hambarde2020s2dnet} are reported in respective literatures. While Wofk et al. \cite{wofk2019fastdepth} only report the values of RMSE and $\delta_1$, we run their released code and models on our desktop to get the values of REL, $\delta_2$, $\delta_3$ and the running time on the GTX 1080 GPU. Table \ref{table: comparison_on_nyu} reports all experimental results. \par 
\begin{figure}[h]
	\centering
	\includegraphics[width=.9\linewidth]{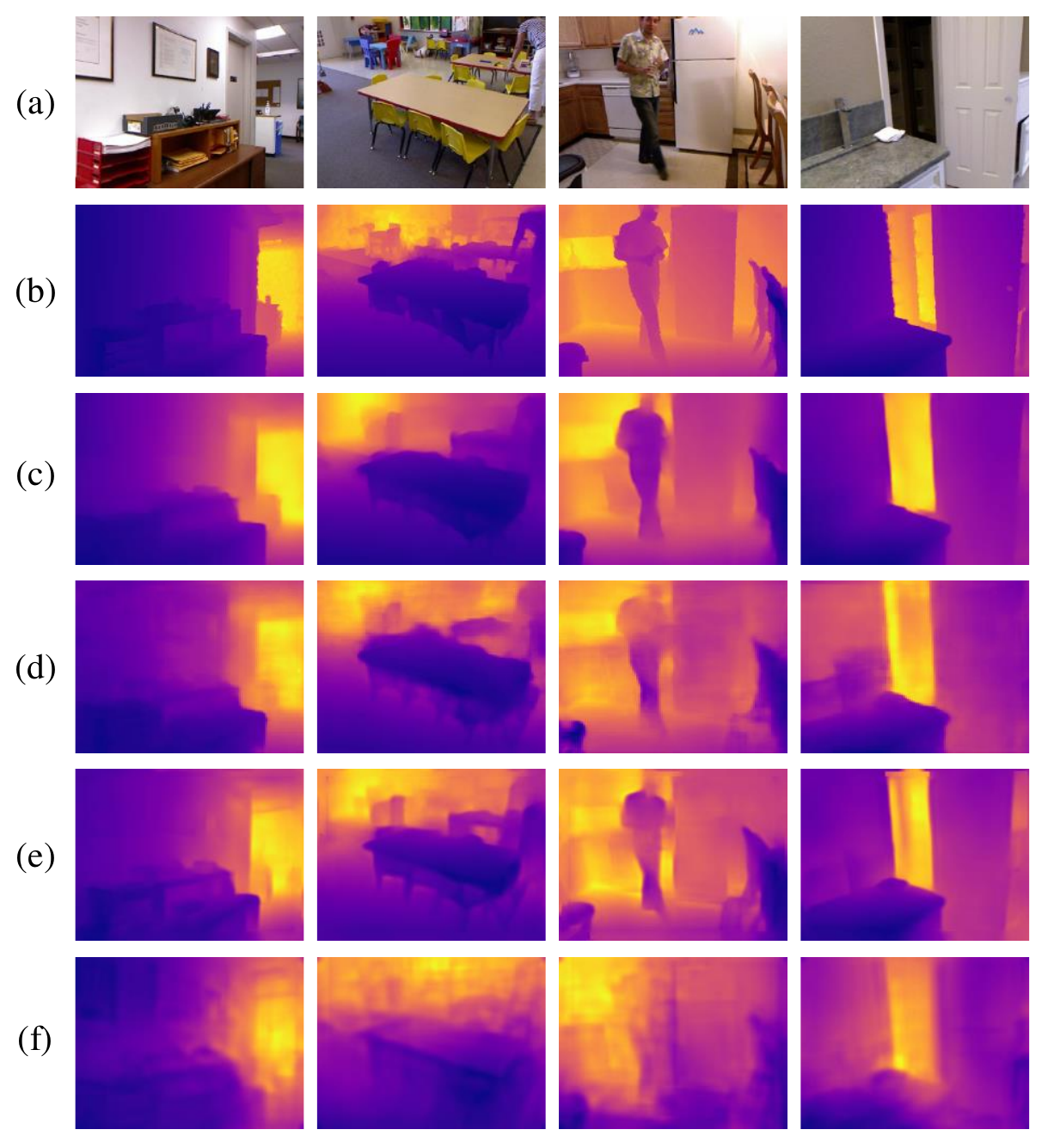} \\
	\caption{Qualitative comparison on the NYU depth v2 dataset. (a) RGB, (b) Groundtruth, (c) Our results, (d) Wofk et al. \cite{wofk2019fastdepth} (Final), (e) Eigen and Fergus \cite{eigen2015predicting}, and (f) Eigen et al. \cite{eigen2014depth} (Fine). Color represents depth (yellow is far, blue is close).}
	\label{fig:comaprison with state-of-the-arts on NYU dataset}
\end{figure}
As can be seen, MobileXNet performs much better than Spek et al. \cite{spek2018cream} in both error and accuracy metrics, and it generates the best RMSE result. Regarding REL, $\delta_1$, $\delta_2$ and $\delta_3$ metrics, MobileXNet also outperforms \cite{eigen2015predicting, liu2015learning, he2018learning, wofk2019fastdepth, yang2019fast, hambarde2020s2dnet}. In addition, the REL, $\delta_1$, $\delta_2$ and $\delta_3$ values of MobileXNet are on par with \cite{laina2016deeper, xu2018structured, cao2017estimating, li2018monocular}. Compared with these four methods, MobileXNet has a shallow and simple architecture, which consists of significantly less layers. Specifically, MobileXNet uses the first 9 layers of MobileXNet as backbone, which is much shallower than the ResNet-50 and ResNet-152 utilized by \cite{laina2016deeper, xu2018structured, li2018monocular, cao2017estimating}. Regarding the decoder network, our upsample layer consists of a $3\times3$ convolutional layer and bilinear interpolation, while Laina et al. \cite{laina2016deeper} (UpProj) utilizes a relatively complex Up-projection method. In addition, \cite{li2018monocular} and \cite{xu2018structured} integrate multi-scale side output feature maps and \cite{cao2017estimating} uses CRF as post-processing operation. \par
\begin{table*}[!ht]
\renewcommand{\arraystretch}{1.3}
	\caption{Comparison of performances on the KITTI Eigen-split \cite{eigen2014depth}. {$\diamondsuit$} means the network does not use CNN designed for image classification, {$\triangle$} represents the first 9 layers of MobileNet. The \textcolor{red}{red} and \textcolor{red}{\textbf{bold}} values indicate the best results.}
	\label{table: comparison_on_kitti_raw}
	\centering
	\begin{tabular}{ c | c | c | c | c | c | c | c | c | c | c } 
		\hline
		Row & Method & Cap & Backbone & \cellcolor{gray} \makecell{RMSE} & \cellcolor{gray} \makecell{REL} & {\cellcolor{pink} \makecell{${{\delta} _1}$}} & {\cellcolor{pink} \makecell{${{\delta}_2}$}} & {\cellcolor{pink} \makecell{${{\delta}_3}$}} & \cellcolor{gray} \makecell{$t_{GPU}$}  & Device \\ 
		\hline\hline
		1 & Eigen et al. \cite{eigen2014depth} & 80 m & $\diamondsuit$ & 7.156 & 0.190 & 0.692 & 0.899 & 0.967 & N/A & N/A \\ 
		2 & Liu et al. \cite{liu2015learning} & 80 m & VGG-16 & 6.986 & 0.217 & 0.647 & 0.882 & 0.961 & N/A & N/A \\  
		3 & Cao et al. \cite{cao2017estimating} & 80 m & ResNet-152 & 4.712 & 0.115 & \textcolor{red}{\textbf{0.887}} & \textcolor{red}{\textbf{0.963}} & 0.982 & N/A & N/A \\
		4 & Godard et al. \cite{godard2017unsupervised} & 80 m & VGG-16 & 5.927 & 0.148 & 0.803 & 0.922 & 0.964 & 35 ms & \makecell{Nvidia Titan-X} \\
        5 & Zhou et al. \cite{zhou2017unsupervised} & 80 m & $\diamondsuit$ & 6.856 & 0.208 & 0.678 & 0.885 & 0.957 & 30 ms & Nvidia Titan-X \\
		6 & Li et al. \cite{li2018monocular} & 80 m & ResNet-50 & 5.325 & 0.128 & 0.833 & 0.956 & \textcolor{red}{\textbf{0.985}} & N/A & N/A \\			 
		7 & Poggi et al. \cite{poggi2018towards} & 80 m & $\diamondsuit$ & 5.907 & 0.146 & 0.801 & 0.926 & 0.967 & 20 ms & Nvidia Titan-X \\
		8 & Casser et al. \cite{casser2019depth} & 80 m & $\diamondsuit$ & 5.521 & 0.142 & 0.820 & 0.942 & 0.976 & 34 ms & Nvidia GTX 1080Ti \\
		9 & Yusiong and Naval \cite{yusiong2019asianet} & 80 m & $\diamondsuit$ & 5.909 & 0.145 & 0.824 & 0.936 & 0.970 & 160 ms & Nvidia GTX 1080Ti \\
		10 & Eom et al. \cite{eom2020temporally} & 80 m & $\diamondsuit$ & \textcolor{red}{\textbf{4.537}} & 0.117 & 0.865 & 0.958 & 0.983 & 130 ms & Nvidia Titan-X \\
		11 & Hambarde and Murala \cite{hambarde2020s2dnet} & 80 m & $\diamondsuit$ & 5.285 & 0.142 & 0.797 & 0.932 & 0.975 & N/A & N/A \\				
		12 & Liu et al. \cite{liu2020mininet} & 80 m & $\diamondsuit$ & 5.264 & 0.141 & 0.825 & 0.941 & 0.976 & 18.57 ms & Nvidia GTX 1080Ti \\
		13 & Ye et al. \cite{ye2020dpnet} & 80 m & ResNet-101 & 4.978 & 0.112 & 0.842 & 0.947 & 0.973 & N/A & N/A \\		
		14 & MobileXNet (Ours) & 80 m & $\triangle$ & 4.965 & \textcolor{red}{\textbf{0.103}} & 0.873 & 0.959 & \textcolor{red}{\textbf{0.985}} & \textcolor{red}{\textbf{15.4 ms}} & Nvidia GTX 1080 \\ \hline \hline
		15 & Cao et al. \cite{cao2017estimating} & 50 m & ResNet-152 & 3.605 & 0.107 & \textcolor{red}{\textbf{0.898}} & \textcolor{red}{\textbf{0.966}} & 0.984 & N/A & N/A \\
		16 & Godard et al. \cite{godard2017unsupervised} & 50 m & VGG-16 & 4.471 & 0.140 & 0.818 & 0.931 & 0.969 & 35 ms & \makecell{Nvidia Titan-X} \\
		17 & Zhou et al. \cite{zhou2017unsupervised} & 50 m & $\diamondsuit$ & 5.181 & 0.201 & 0.696 & 0.900 & 0.966 & 30 ms & Nvidia Titan-X \\		 		
		18 & Poggi et al. \cite{poggi2018towards} & 50 m & $\diamondsuit$ & 4.608 & 0.145 & 0.813 & 0.934 & 0.972 & 20 ms & Nvidia Titan-X \\
		19 & Yusiong and Naval \cite{yusiong2019asianet} & 50 m & $\diamondsuit$ & 4.014 & 0.122 & 0.864 & 0.953 & 0.978 & 160 ms & Nvidia GTX 1080Ti \\
		20 & Eom et al. \cite{eom2020temporally} & 50 m & $\diamondsuit$ & \textcolor{red}{\textbf{3.493}} & 0.113 & 0.877 & 0.963 & 0.985 & 130 ms & Nvidia Titan-X \\ 		
		21 & Liu et al. \cite{liu2020mininet} & 50 m & $\diamondsuit$ & 4.067 & 0.135 & 0.838 & 0.947 & 0.978 & 18.57 ms & Nvidia GTX 1080Ti \\
        22 & Spek et al. \cite{spek2018cream} & 50 m & ERFNet & 4.363 & 0.156 & 0.818 & 0.940 & 0.977 & \textcolor{red}{\textbf{3.2 ms}} & Nvidia GTX 1080Ti \\
		23 & MobileXNet (Ours) & 50 m & $\triangle$ & 3.842 & \textcolor{red}{\textbf{0.098}} & 0.886 & \textcolor{red}{\textbf{0.966}} & \textcolor{red}{\textbf{0.987}} & 15.4 ms & Nvidia GTX 1080 \\ \hline
	\end{tabular}
	
	\begin{tabular}{| c | c | c |} \hline
		\cellcolor{gray}Lower is better & \cellcolor{pink}Higher is better & {$\delta_1: \delta < 1.25$}, {$\delta_2: \delta < 1.25^{2}$}, {$\delta_3: \delta < 1.25^{3}$} \\ \hline 
	\end{tabular} 	
\end{table*}
It can be observed from Table \ref{table: comparison_on_nyu} that Spek et al. \cite{spek2018cream} yields the fastest running speed, while it performs worse than most of the counterparts in terms of the accuracy and error metrics. Our MobileXNet runs slightly slower than Wofk et al. \cite{wofk2019fastdepth}, which is tested at $224\times224$ sized images. It should be noted that the MobileXNet runs on images having $228\times304$ pixels. The running speed of MobileXNet is significantly faster (7.9ms vs 150ms) than Xu et al. \cite{xu2018structured}, which was tested on a Titan-X GPU. While the running time is obtained with different GPUs, the Titan-X GPU has 3584 CUDA cores and 12GB memory, which is much stronger than our 1080 GPU (2580 CUDA cores and 8GB memory). With the same sized images, the speed of MobileXNet is about 7$\times$ to 10$\times$ faster than \cite{laina2016deeper}. It should be noted that \cite{laina2016deeper} was evaluated on a GTX Titan GPU which has 3072 CUDA cores and 12GB memory. 

\begin{figure*}[h]
	\centering
	\subfloat[]{
		\includegraphics[width=.25\linewidth]{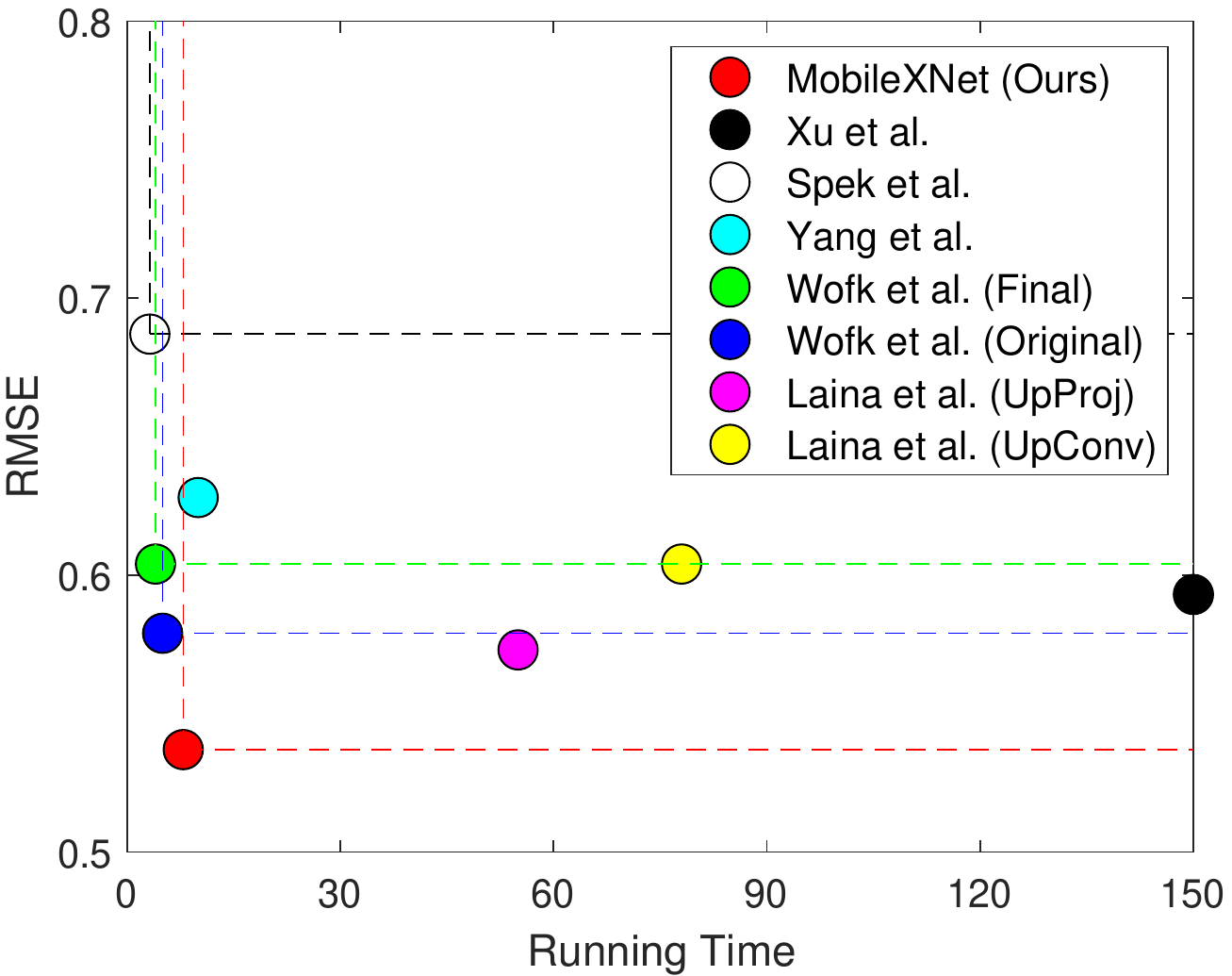}
		\label{fig:pareto_rmse_nyu}}
	\subfloat[]{
		\includegraphics[width=.25\linewidth]{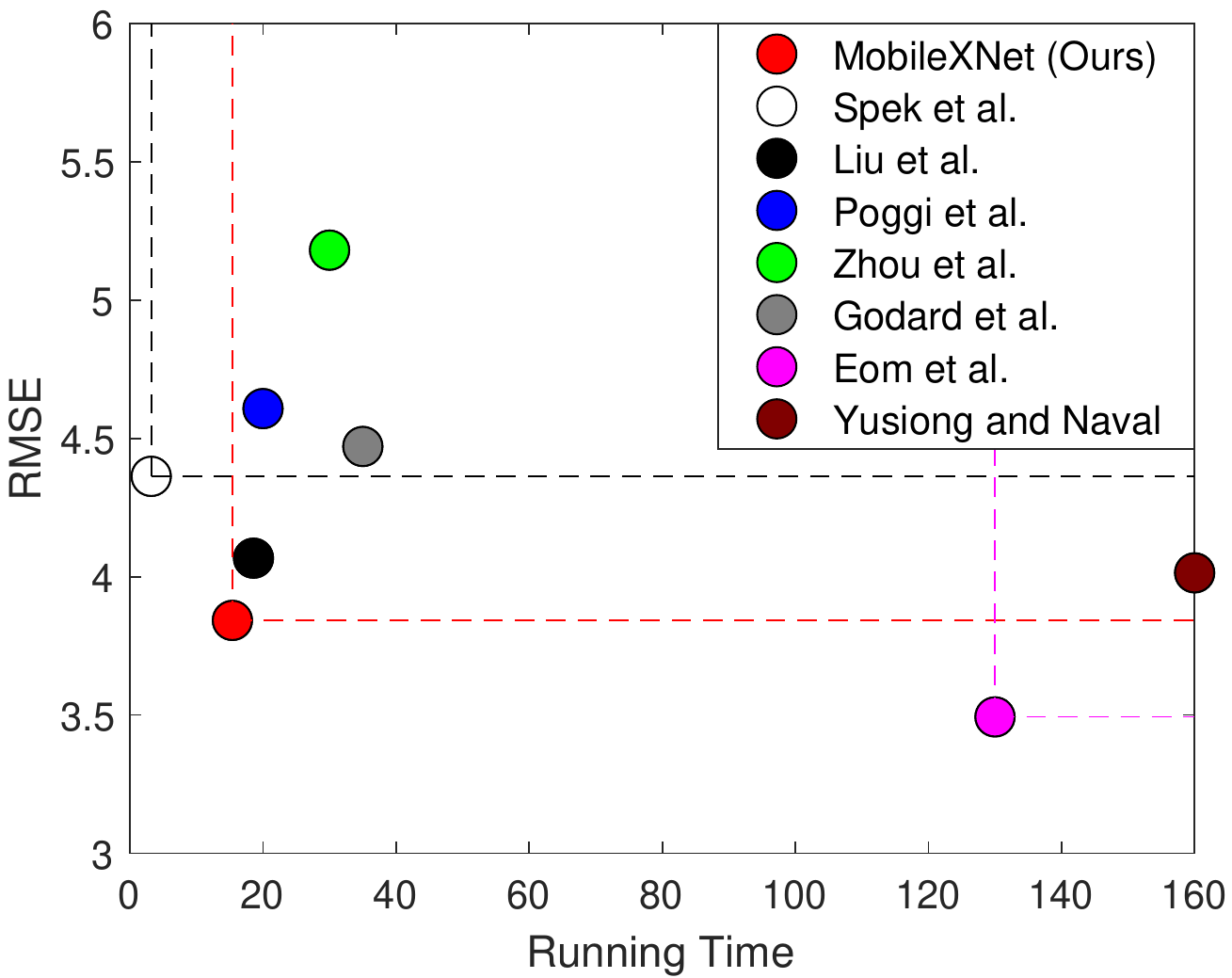}
		\label{fig:pareto_rmse_kitti_cap50}}
	\subfloat[]{
		\includegraphics[width=.25\linewidth]{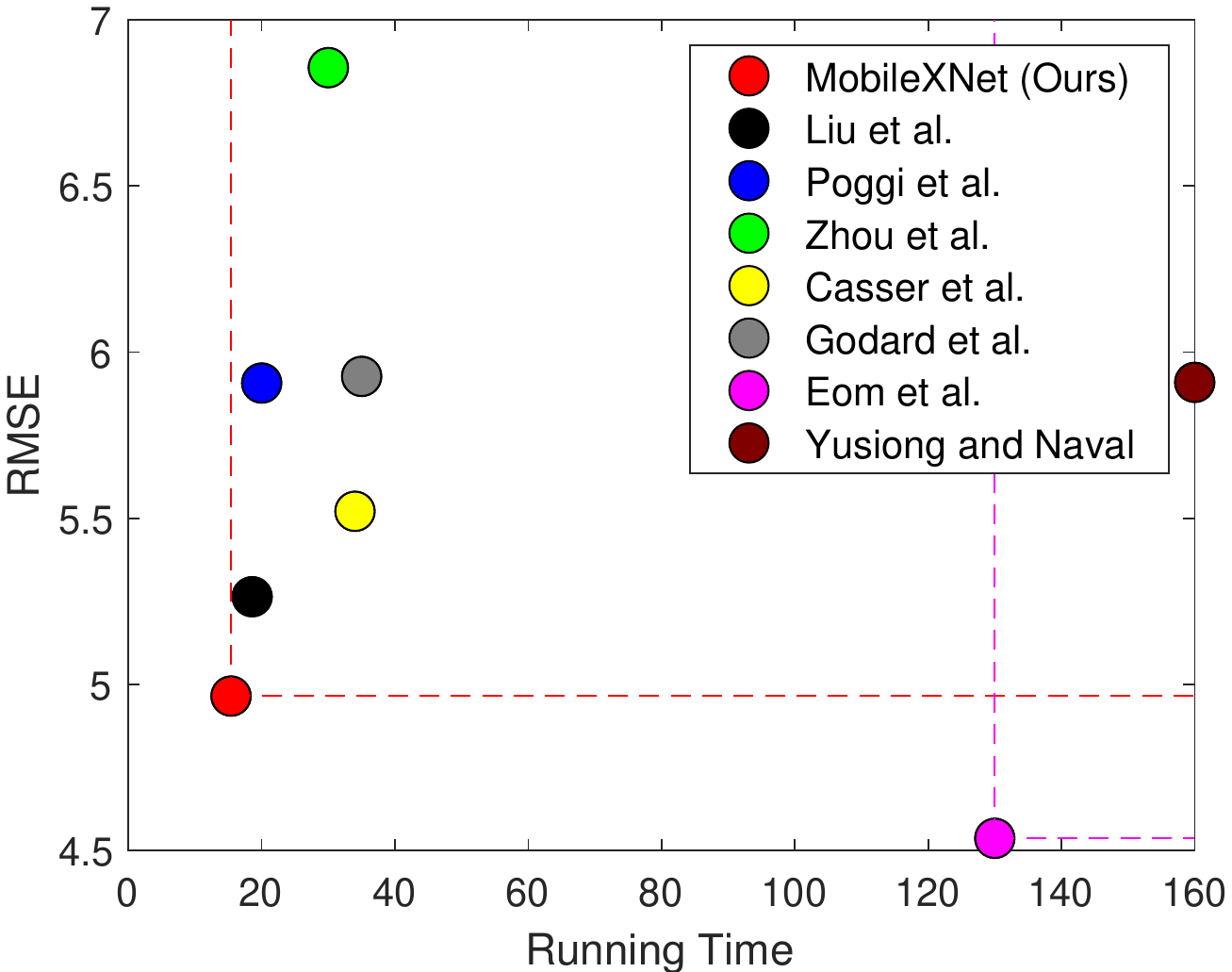}
		\label{fig:pareto_rmse_kitti_cap80}}
	\caption{Pareto Optimality. (a) The NYU depth v2 dataset, (b) The KITTI dataset (range caps of 50m), and (c) The KITTI dataset (range caps of 80m).}
	\vspace{-0.1in}
\end{figure*}

The non-dominated algorithms over running time and RMSE are shown in Fig. \ref{fig:pareto_rmse_nyu}. As can be seen, MobileXNet dominates Laina et al. \cite{laina2016deeper} (UpConv), Laina et al. \cite{laina2016deeper} (UpProj), Xu et al. \cite{xu2018structured} and Yang et al. \cite{yang2019fast}. Besides, MobileXNet, Wofk et al. \cite{wofk2019fastdepth} (Original), Wofk et al. \cite{wofk2019fastdepth} (Final) and Spek et al. \cite{spek2018cream} lie on the Pareto Front because none of them is dominated by another. Although MobileXNet is slower than \cite{wofk2019fastdepth} (Original), \cite{wofk2019fastdepth} (Final) and Spek et al. \cite{spek2018cream}, its accuracy is much better than them. More importantly, MobileXNet runs about 126 fps on a less powerful GPU, which is adequate for the real-time application of autonomous driving and robotics. Considering the trade-off between accuracy and speed, MobileXNet is the best compromise solution. In Fig. \ref{fig:comaprison with state-of-the-arts on NYU dataset}, we provide qualitative comparison of the proposed method with \cite{eigen2014depth, eigen2015predicting, wofk2019fastdepth}. It is clearly observed that the results of \cite{eigen2014depth, eigen2015predicting, wofk2019fastdepth} are blurry. By contrast, our method recovers more details and the predicted depth maps are much clearer than its counterparts. \par 
\subsection{KITTI Dataset}
The KITTI dataset \cite{geiger2013vision} consists of outdoor scene images with a resolution of $375\times1241$ pixels. This dataset has sparse depth images captured by a Velodyne LiDAR. We utilize the same split as Eigen et al. \cite{eigen2014depth}, where only left images, which includes 22600 training images and 697 testing images are used. To generate the groundtruth depth maps, we projected corresponding Velodyne data points to the left image plane. The missed depth values in the groundtruth depth maps are ignored both in training and testing. As the LiDAR provides no measurements in the upper part of the images, only the bottom $228\times912$ pixels region is used in this study. \par
In order to compare with state-of-the-art methods \cite{eigen2014depth, liu2015learning, zhou2017unsupervised, godard2017unsupervised, cao2017estimating, li2018monocular, poggi2018towards, spek2018cream, casser2019depth, eom2020temporally, yusiong2019asianet, ye2020dpnet, liu2020mininet}, we evaluate our method with the depth ranging from 0m to 80m and 0m to 50m. Table \ref{table: comparison_on_kitti_raw} shows the results of our method together with baselines. For the depth range from 0m to 80m, MobileXNet achieves the best REL and $\delta_3$ results, while the value of RMSE is comparable to \cite{eom2020temporally, cao2017estimating}. Moreover, the $\delta_1$ and $\delta_2$ of MobileXNet are only slightly inferior to Cao et al. \cite{cao2017estimating}. It should be noted that \cite{cao2017estimating} is built on top of an extremely deep CNN, ResNet-152, and uses fully connected CRF for post-processing. In addition, Eom et al. \cite{eom2020temporally} adopts a two-stream encoder in order to learn features from RGB image and optical flow. However, MobileXNet does not exploit any multi-stream architecture or post-possessing operation, and has much less layers than \cite{cao2017estimating}. In addition, our method outperforms \cite{eigen2014depth, zhou2017unsupervised, godard2017unsupervised, li2018monocular, poggi2018towards, casser2019depth, yusiong2019asianet, ye2020dpnet, liu2020mininet}. For the depth range of 0 to 50 meters, MobileXNet produced the best performance on REL, $\delta_2$, and $\delta_3$, and the second best performance of $\delta_1$. It demonstrates that MobileXNet works better for close-range depth. \par
\begin{figure}[h]
	\centering
	\includegraphics[width=.9\linewidth]{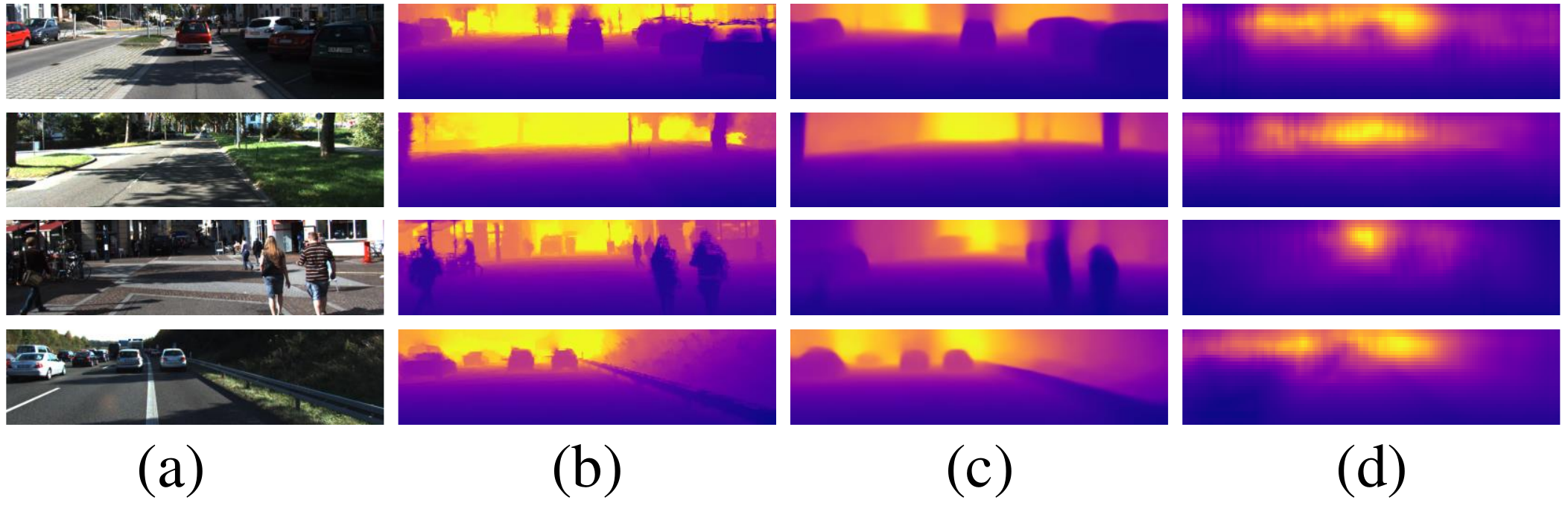} \\
	\caption{Qualitative comparison on the KITTI dataset. (a) RGB, (b) Groundtruth, (c) Our results, and (d) Eigen et al. \cite{eigen2014depth} (Fine). The groundtruth has been interpolated for visualization. Color represents depth (yellow is far, blue is close).}
	\label{fig:comaprison with state-of-the-art on KITTI dataset}
\end{figure}
In terms of running speed, our MobileXNet only needs 15.4ms to process a $228\times912$ sized image on a GTX 1080 GPU, which is faster than its counterparts such as \cite{godard2017unsupervised, zhou2017unsupervised, poggi2018towards, casser2019depth, yusiong2019asianet, eom2020temporally, liu2020mininet}. It is worth noting that \cite{godard2017unsupervised, zhou2017unsupervised, poggi2018towards, casser2019depth, yusiong2019asianet, eom2020temporally, liu2020mininet} were tested on GPUs which are much stronger than ours. Specifically, the Titan-X GPU has 3584 CUDA cores and 12GB memory, the GTX 1080Ti GPU has 3584 CUDA cores and 11GB memory, while our GTX 1080 GPU only has 2560 CUDA cores and 8GB memory. \par 
We use non-dominance to compare the running time and RMSE of the proposed MobileXNet, \cite{godard2017unsupervised, zhou2017unsupervised, poggi2018towards, spek2018cream, casser2019depth, eom2020temporally, yusiong2019asianet, liu2020mininet}. It can be observed from Fig. \ref{fig:pareto_rmse_kitti_cap50} and Fig. \ref{fig:pareto_rmse_kitti_cap80} that MobileXNet and Eom et al. \cite{eom2020temporally} lie on the Pareto Front in the depth range between 0m and 80m. However, the MobileXNet is more than 8$\times$ faster than Eom et al. \cite{eom2020temporally}, even though \cite{eom2020temporally} was tested on a more powerful Titan-X GPU. As can be observed from Table \ref{table: comparison_on_kitti_raw}, MobileXNet also outperforms Eom et al. \cite{eom2020temporally} in REL, $\delta_1$, $\delta_2$, and $\delta_3$ for depth range from 0m to 80m as well as 0m to 50m. According to the non-dominated set and Pareto front distribution, MobileXNet is the best solution among these methods. For the depth range of 0 to 50m, MobileXNet and Spek et al. \cite{spek2018cream} are none-dominated by each other. According to Table VI, the MobileXNet performs much better than Spek et al. \cite{spek2018cream} with respect to the error and accuracy metrics. Moreover, the MobileXNet achieves a real-time speed of 65 fps at a larger resolution (228 × 912 pixels). Hence, MobileXNet is the best solution among these methods. Qualitative results are shown in Fig. \ref{fig:comaprison with state-of-the-art on KITTI dataset}. As can be observed, our method generates much clearer depth maps than \cite{eigen2014depth}. \par
\begin{table*}[h]
\renewcommand{\arraystretch}{1.3}
	\caption{Comparison of performances on $93.5\%$ of the KITTI Eigen-split with accurate groundtruth labels released by the KITTI evaluation benchmark. K denotes the KITTI dataset, CS refers to the Cityscapes dataset, CS + K represents training on the Cityscapes dataset then fine-tuning on the KITTI dataset. {$\triangle$} represents the first 9 layers of MobileNet. The \textcolor{red}{red} and \textcolor{red}{\textbf{bold}} values indicate the best results.}
	\centering
	\label{table: comparison_on_kitti_official}
	\begin{tabular}{ c | c | c | c | c | c | c | c | c | c } 
		\hline 
		Row & Method & Cap & Backbone & Dataset & \cellcolor{gray} \makecell{RMSE} & \cellcolor{gray} \makecell{REL} & {\cellcolor{pink} \makecell{${{\delta}_1}$}} & {\cellcolor{pink} \makecell{${{\delta}_2}$}} & {\cellcolor{pink} \makecell{${{\delta}_3}$}} \\ [0.5ex] 
		\hline\hline
		1 & Amiri et al. \cite{amiri2019semi} & 80 m & ResNet-50 & {K} & \textcolor{red}{\textbf{3.995}} & 0.096 & 0.892 & 0.972 & 0.992 \\
		2 & Godard et al. \cite{godard2017unsupervised} & 80 m & ResNet-50 & {CS+K} & 4.279 & 0.097 & 0.898 & 0.973 & 0.991 \\
		3 & Aleotti et al. \cite{aleotti2018generative} & 80 m & VGG-16 & {CS+K} & 4.236 & 0.096 & 0.899 & 0.974 & 0.992 \\ 	
		4 & MobileXNet (Ours) & 80 m & $\triangle$ & {K} & 4.128 & \textcolor{red}{\textbf{0.087}} & \textcolor{red}{\textbf{0.905}} & \textcolor{red}{\textbf{0.976}} & \textcolor{red}{\textbf{0.993}} \\ \hline \hline  
		5 & Godard et al. \cite{godard2017unsupervised} & 50 m & ResNet-50 & {CS+K} & 4.100 & 0.095 & 0.896 & 0.975 & 0.992 \\
		6 & Aleotti et al. \cite{aleotti2018generative} & 50 m & VGG-16 & {CS+K} & 4.110 & 0.094 & 0.897 & 0.976 & 0.993 \\ 
		7 & MobileXNet (Ours) & 50 m & $\triangle$ & {K} & \textcolor{red}{\textbf{3.081}} & \textcolor{red}{\textbf{0.083}} & \textcolor{red}{\textbf{0.916}} & \textcolor{red}{\textbf{0.981}} & \textcolor{red}{\textbf{0.994}} \\ \hline
	\end{tabular}
	
	\begin{tabular}{| c | c | c |} 
		\cellcolor{gray}Lower is better & \cellcolor{pink}Higher is better & {$\delta_1: \delta < 1.25$}, {$\delta_2: \delta < 1.25^{2}$}, {$\delta_3: \delta < 1.25^{3}$} \\ \hline 
	\end{tabular} 
	
\end{table*}
According to \cite{godard2017unsupervised, uhrig2017sparsity}, the depth measurements from the LiDAR may be influenced by many factors such as rotation of the LiDAR and incorrect depth readings resulting from the occlusion at object boundaries. In order to better validate the performance of our proposed MobileXNet, we applied the same split of images with more accurate groundtruth labels provided by the KITTI official \cite{uhrig2017sparsity}. The annotated depth maps are generated by filtering LiDAR points with a computed disparity map from the Semi-Global Matching algorithm \cite{hirschmuller2007stereo} to remove outliers from the raw measurements. It is worth noting that the annotated depth maps are not available for 315 training and 45 testing images of the original Eigen split, thus, the number of training and testing images are 22\,285 and 652 respectively. The missed values in the annotated depth maps are ignored during both training and evaluation. \par
We compare our method with \cite{godard2017unsupervised, aleotti2018generative, amiri2019semi}. The results of \cite{godard2017unsupervised} and \cite{aleotti2018generative} are reported in \cite{aleotti2018generative}. Quantitative results are listed in Table \ref{table: comparison_on_kitti_official}. As shown in rows 1 to 4 of Table \ref{table: comparison_on_kitti_official}, for the cap of 0-80m, MobileXNet outperforms all baselines in terms of REL, $\delta_1$, $\delta_2$ and $\delta_3$, while the RMSE is slightly inferior to \cite{amiri2019semi}. It is worth noting that \cite{amiri2019semi} applies ResNet-50 as encoder, which is much deeper than the network of MobileXNet. Since \cite{amiri2019semi} only reports results in the range of 0-80m, thus, we compare MobileXNet with \cite{godard2017unsupervised} and \cite{aleotti2018generative} for the cap of 0-50m. It can be seen from rows 5 to 7 of Table \ref{table: comparison_on_kitti_official}, with a simple and shallow network architecture the proposed MobileXNet outperforms \cite{godard2017unsupervised} and \cite{aleotti2018generative} in all metrics. \par
\subsection{Make 3D Dataset}
In this part, we evaluate MobileXNet on the Make3D dataset \cite{saxena2008make3d}, consisting of 400 training and 134 testing images. The RGB images have a resolution of $2272\times1704$ pixels, and the corresponding groundtruth depth map is $55\times305$ sized. Following \cite{fu2018deep} all images are downsampled to $568\times426$ pixels. In this study, we train our network on the $460\times345$ pixels center cropped region. To compare with the state-of-the-art, we evaluate the trained model with \textit{C1} (depth range from 0m to 70m) and \textit{C2} (depth range from 0m to 80m) criterions as introduced in \cite{liu2014discrete}. \par
\begin{table}[h]
	\renewcommand{\arraystretch}{1.3}
	\renewcommand{\tabcolsep}{1mm}
	\caption{Qualitative results on the Make3D dataset. The \textcolor{red}{red} and \textcolor{red}{\textbf{bold}} values indicate the best results.}
	\centering
	\label{table: comparison_on_make3d}
	\begin{tabular}{ c | c | c | c | c | c | c }
		\hline
		\multirow{2}*{Method} & \multicolumn{3}{c|}{C1 error}  
		& \multicolumn{2}{c}{C2 error} \\ \cline{2-7}
		& \cellcolor{gray}\makecell{REL} & \cellcolor{gray}\makecell{log10} & \cellcolor{gray}\makecell{RMSE} & \cellcolor{gray}\makecell{REL} & \cellcolor{gray}\makecell{log10} & \cellcolor{gray}\makecell{RMSE} \\ \hline \hline
		Liu et al. \cite{liu2014discrete} & 0.335 & 0.137 & 9.49 & 0.338 & 0.134 & 12.60 \\ 
		Karsch et al. \cite{karsch2014depth} & 0.355 & 0.127 & 9.20 & 0.361 & 0.148 & 15.10 \\ 
		Li et al. \cite{li2015depth} & 0.278 & 0.092 & 7.12 & 0.279 & 0.102 & 10.27 \\ 
		Liu et al. \cite{liu2015learning} & 0.287 & 0.109 & 7.36 & 0.287 & 0.122 & 14.09 \\ 
		Roy and Todorovic \cite{roy2016monocular} & N/A & N/A & N/A & 0.260 & 0.119 & 12.40 \\ 
		Godard et al. \cite{godard2017unsupervised} & 0.443 & 0.143 & 8.860 & N/A & N/A & N/A \\ 
		Kuznietsov et al. \cite{kuznietsov2017semi} & 0.421 & 0.190 & 8.24 & N/A & N/A & N/A \\ 
		Fu et al. \cite{fu2018deep} & 0.236 & \textcolor{red}{\textbf{0.082}} & 7.02 & 0.238 & \textcolor{red}{\textbf{0.087}} & 10.01 \\ 
		Zhao et al. \cite{zhao2019geometry} & 0.403 & N/A & 10.424 & N/A & N/A & N/A \\ 
		MobileXNet (Ours) & \textcolor{red}{\textbf{0.229}} & 0.086 & \textcolor{red}{\textbf{6.807}} & \textcolor{red}{\textbf{0.233}} & 0.089 & \textcolor{red}{\textbf{8.020}} \\ 
		\hline 

	\end{tabular}    
	\begin{tabular}{| c |} 
		{\cellcolor{gray}Lower is better} \\ \hline
	\end{tabular} 
	\label{tab:test on make3d}
\end{table}
\begin{figure}[h]
	\centering
	\includegraphics[width=.9\linewidth]{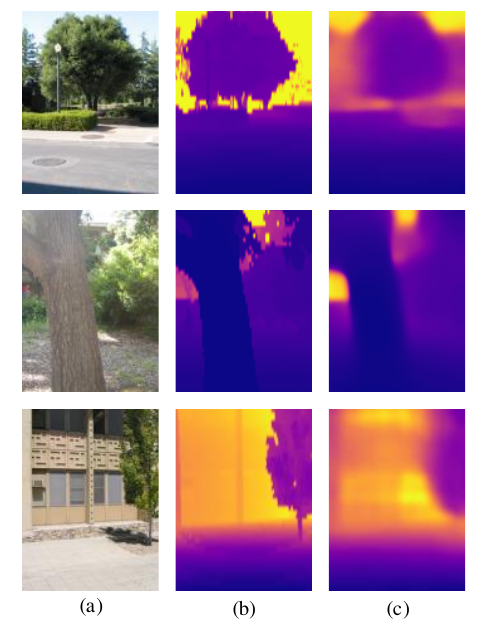} \\
	\caption{Quantitative results on the Make3D dataset. (a) RGB, (b) Groundtruth, and (c) Our results. Color represents depth (yellow is far, blue is close).}
	\label{fig:qualitative results on Make3d dataset}
\end{figure}
The previous methods \cite{liu2014discrete, karsch2014depth, liu2015learning, li2015depth, roy2016monocular, godard2017unsupervised, kuznietsov2017semi, fu2018deep, zhao2019geometry} are used as baseline, all results are shown in Table \ref{table: comparison_on_make3d}. As can be observed, MobileXNet outperforms \cite{liu2014discrete, karsch2014depth, liu2015learning, li2015depth, roy2016monocular, godard2017unsupervised, kuznietsov2017semi, zhao2019geometry} in all metrics. As for \cite{fu2018deep}, our method generates better REL and RMSE results, while the log10 value is slightly inferior. It is worth noting that \cite{fu2018deep} applies a deep CNN as backbone and combines a scene understanding module. In addition, our MobileXNet only needs 0.012s to compute a depth map on a single GTX 1080 GPU. For Liu et al. \cite{liu2015learning}, 1.1s is required to infer depth map from a test image. However, the time for computing superpixels is not included. Qualitative results from this dataset are shown in Fig. \ref{fig:qualitative results on Make3d dataset}. \par 
\subsection{UnrealDataset}
To demonstrate the ability of predicting depth maps from small sized images, we evaluate the proposed MobileXNet on the UnrealDataset \cite{mancini2018j}. This data set contains over 100K images collected in a series of simulated urban and forest scenarios, and the groundtruth depth up to 40 meters. Compared with the aforementioned data sets \cite{silberman2012indoor, geiger2013vision, saxena2008make3d}, the RGB and depth images in the UnrealDataset have a much smaller size, $160\times256$ pixels. The Unreal Engine and AirSim softwares simulate an MAV flying in the simulated environments. The RGB and depth image pairs are collected from the MAV's frontal camera using a plugin. We first evaluate the proposed method on original images, and compare it with \cite{ronneberger2015u, mancini2018j, ma2018sparse, wofk2019fastdepth}. Considering the limitation of GPU memory, we choose the ResNet50-DeConv3{\footnote{The fully convolutional part of ResNet-50 is used as the encoder and the deconvolution with $3\times3$ kernel is applied for upsampling.}} combination of \cite{ma2018sparse}. We train \cite{ma2018sparse, wofk2019fastdepth} with the official released source code and default parameters. As in section V, we train the Pytorch implemented U-Net (Bilinear) with the hybrid loss function. \par
\begin{figure}[h]
	\centering
	\includegraphics[width=.9\linewidth]{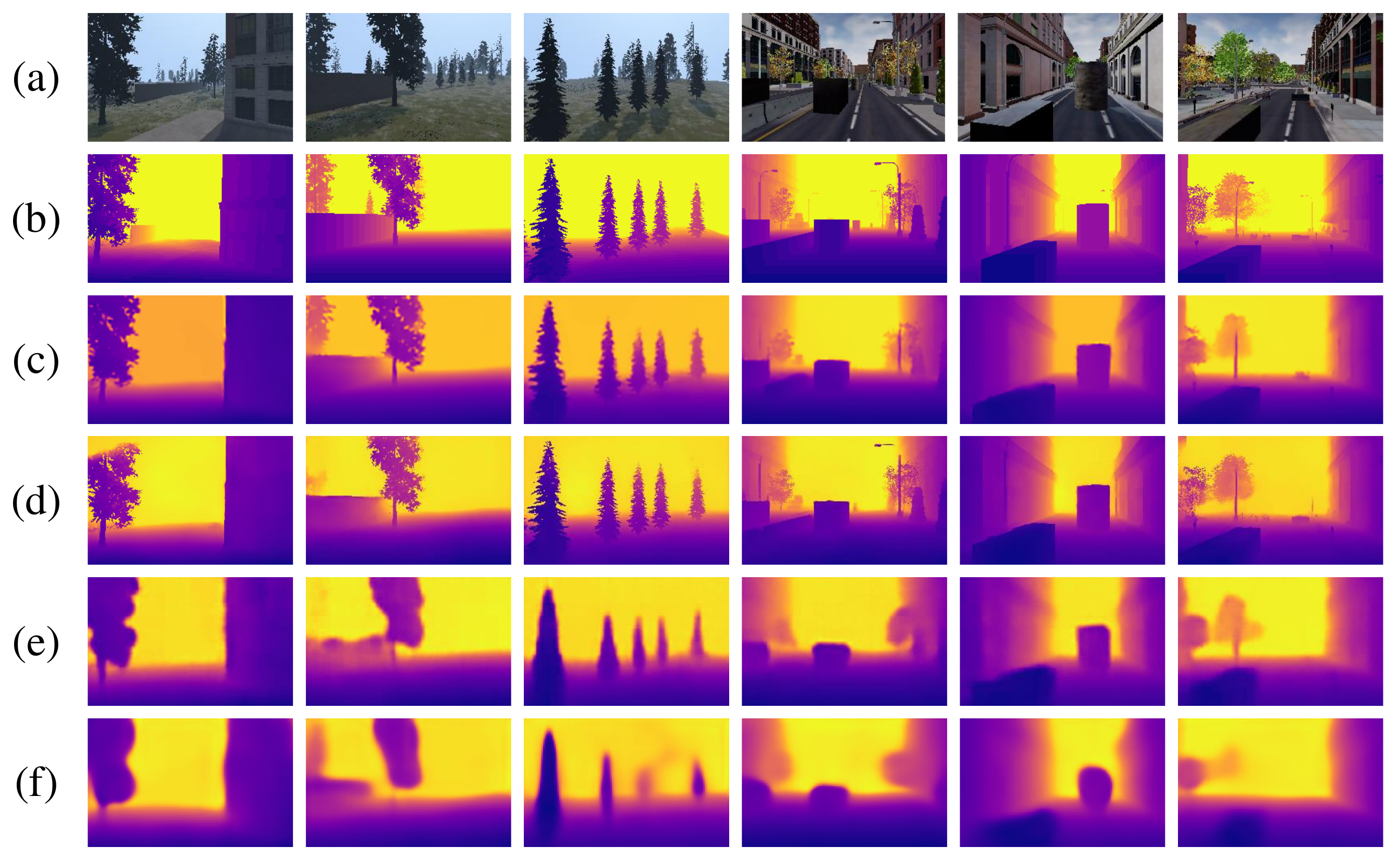} \\
	\caption{Qualitative comparison on the UnrealDataset (original). (a) RGB, (b) Groundtruth, (c) Our results, (d) U-Net \cite{ronneberger2015u}, (e) Ma and Karaman \cite{ma2018sparse}, and (f) Wofk et al. \cite{wofk2019fastdepth} (Original). Color represents depth (yellow is far, blue is close).}
	\label{fig: qualitive comaprison on the original images of Unreal dataset}
\end{figure}

From rows 1 to 5 of Table \ref{table: comparison_on_unreal}, we show the quantitative comparison between MobileXNet and baselines on the original images. Following \cite{mancini2018j}, we only report the RMSE and running time of J-MOD2. As can be observed, MobileXNet outperforms \cite{ma2018sparse, wofk2019fastdepth} in all metrics. It performs best in REL, $\delta_3$ and speed, while the RMSE value of MobileXNet is slightly inferior to J-MOD2. J-MOD2 is a multi-task learning method, which jointly learns object detection and depth estimation. The object detection branch informs the depth estimation branch with object structures, which improves the accuracy of depth estimation. However, our method only works out MDE. According to Fig. \ref{fig:pareto_rmse_unreal}a, both MobileXNet and J-MOD2 are non-dominated solutions. The running time of J-MOD2 is longer than MobileXNet, even though J-MOD2 was evaluated on a more powerful GPU. The threshold metric of MobileXNet is on par with U-Net \cite{ronneberger2015u}. It is worth noting that images from the UnrealDataset are much smaller than the NYU v2 dataset \cite{silberman2012indoor}. As both U-Net and MobileXNet downsample the feature maps to 1/16 scale of the input images, the produced feature maps include more spatial information than \cite{ma2018sparse, wofk2019fastdepth}. In addition, U-Net \cite{ronneberger2015u} is twice as wide as our MobileXNet, which enables U-Net to learn more information. Some qualitative results of our method and baselines are shown in Fig. \ref{fig: qualitive comaprison on the original images of Unreal dataset}, which further demonstrates the superior performance of MobileXNet. \par
\begin{figure}[!h]
	\centering
	\includegraphics[width=.9\linewidth]{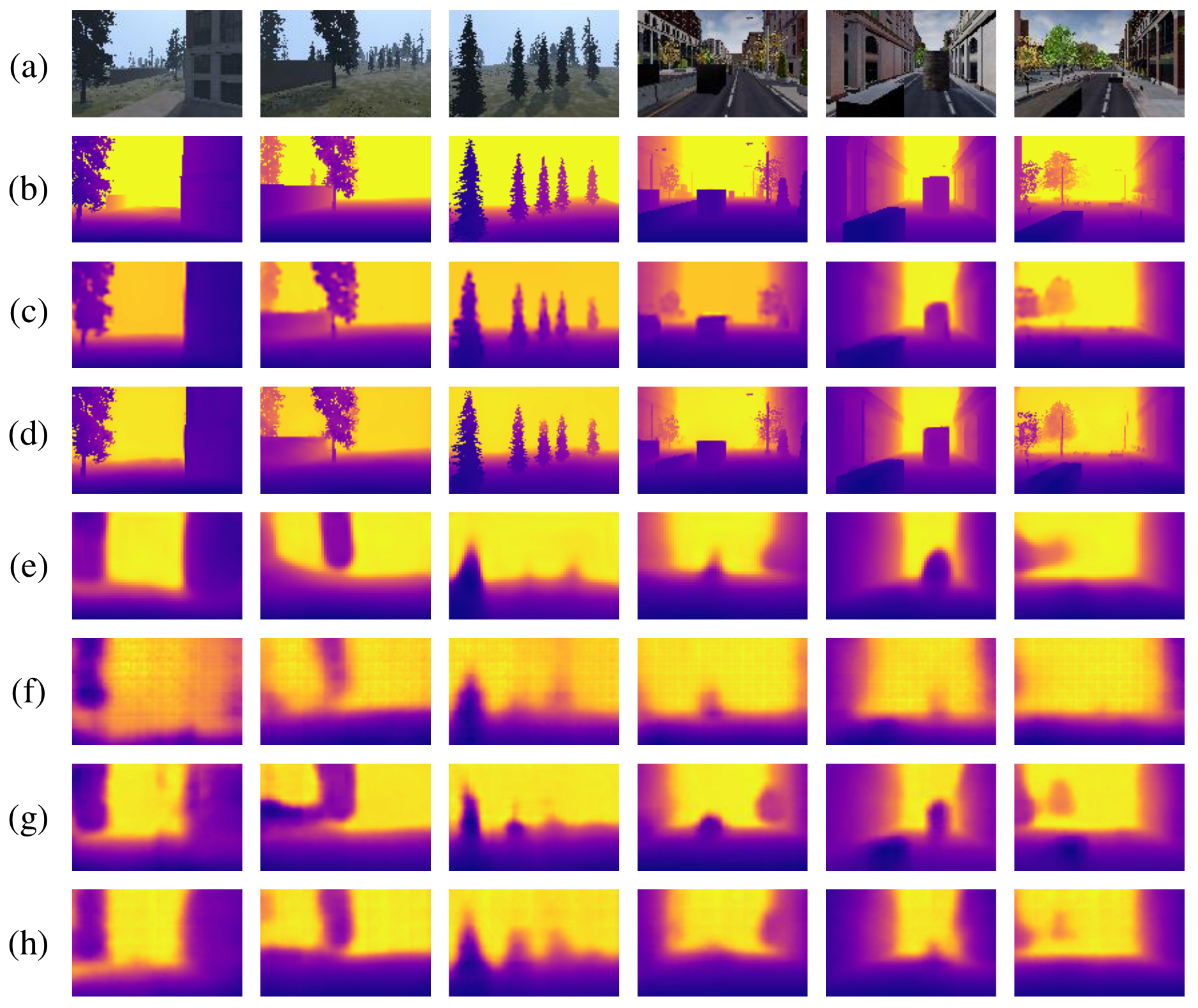} \\
	\caption{Qualitative comparison on the UnrealDataset ($80\times128$). (a) RGB, (b) Groundtruth, (c) Our results, (d) U-Net \cite{ronneberger2015u}, (e) Wofk et al. \cite{wofk2019fastdepth} (Original), (f) Ma and Karaman \cite{ma2018sparse}, (g) Ma and Karaman \cite{ma2018sparse} with $5\times5$ sized filters in the input layer, and (h) Ma and Karaman \cite{ma2018sparse} with $3\times3$ sized filters in the input layer. Color represents depth (yellow is far, blue is close).}
	\label{fig: qualitive comaprison on the half sized images of Unreal dataset}
\end{figure}
\begin{table*}[h]
	\renewcommand{\arraystretch}{1.3}
	\caption{Comparison of performances on the UnrealDataset. {$\diamondsuit$} means the network does not use CNN designed for image classification, $\triangle$ represents the first 9 layers of MobileNet. Ma and Karaman ($3\times3$) and Ma and Karaman ($5\times5$) denote the use of $3\times3$ and $5\times5$ sized filters in the input layer respectively. The \textcolor{red}{red} and \textcolor{red}{\textbf{bold}} values indicate the best results.}
	\centering
	\label{table: comparison_on_unreal}
	\begin{tabular}{ c | c | c | c | c | c | c | c | c | c | c } 
		\hline
		Row & Method & Backbone & Input Size & \cellcolor{gray} \makecell{RMSE} & \cellcolor{gray} \makecell{REL} & {\cellcolor{pink} \makecell{${{\delta}_1}$}} & {\cellcolor{pink} \makecell{${{\delta}_2}$}} & {\cellcolor{pink} \makecell{${{\delta}3}$}} & \cellcolor{gray} \makecell{$t_{GPU}$}  & Device \\
		\hline\hline
		1 & J-MOD2 \cite{mancini2018j} & VGG-19 & $160\times256$ &\textcolor{red}{\textbf{3.473}} & N/A & N/A & N/A & N/A & 10 ms & Nvidia Titan-X \\ 
		2 & U-Net \cite{ronneberger2015u} & $\diamondsuit$ & $160\times256$ & 3.834 & 0.130 & \textcolor{red}{\textbf{0.884}} & \textcolor{red}{\textbf{0.957}} & \textcolor{red}{\textbf{0.977}} & 8.8 ms & Nvidia GTX 1080 \\
		3 & Ma and Karaman \cite{ma2018sparse} & ResNet-50 & $160\times256$ & 4.023 & 0.126 & 0.870 & 0.951 & 0.974 & 6.9 ms & Nvidia GTX 1080 \\		
		4 & Wofk et al. \cite{wofk2019fastdepth} & MobileNet & $160\times256$ & 4.208 & 0.155 & 0.860 & 0.945 & 0.972 & 6.7 ms & Nvidia GTX 1080 \\
		5 & MobileXNet (Ours) & $\triangle$ & $160\times256$ & 3.612 & \textcolor{red}{\textbf{0.124}} & 0.878 & 0.954 & \textcolor{red}{\textbf{0.977}} & \textcolor{red}{\textbf{6.6 ms}} & Nvidia GTX 1080 \\ \hline \hline
		6 & U-Net \cite{ronneberger2015u} & $\diamondsuit$ & $128\times204$ & 3.917 & 0.134 & \textcolor{red}{\textbf{0.879}} & \textcolor{red}{\textbf{0.956}} & 0.969 & 6.9 ms& Nvidia GTX 1080 \\ 
		7 & Ma and Karaman \cite{ma2018sparse} & ResNet-50 & $128\times204$ & 4.188 & 0.138 & 0.856 & 0.946 & 0.973 & 6.6 ms& Nvidia GTX 1080 \\		
		8 & Wofk et al. \cite{wofk2019fastdepth} & MobileNet & $128\times204$ & 4.393 & 0.157 & 0.852 & 0.940 & 0.976 & 6.4 ms & Nvidia GTX 1080 \\ 				
		9 & MobileXNet (Ours) & $\triangle$ & $128\times204$ & \textcolor{red}{\textbf{3.733}} & \textcolor{red}{\textbf{0.133}} & 0.871 & 0.954 & \textcolor{red}{\textbf{0.977}} & \textcolor{red}{\textbf{6.3 ms}} & Nvidia GTX 1080 \\ \hline \hline
		10 & U-Net \cite{ronneberger2015u} & $\diamondsuit$ & $80\times128$ & 3.912 & \textcolor{red}{\textbf{0.135}} & \textcolor{red}{\textbf{0.873}} & \textcolor{red}{\textbf{0.956}} & \textcolor{red}{\textbf{0.977}} & \textcolor{red}{\textbf{4.8 ms}} & Nvidia GTX 1080 \\ 
		11 & Ma and Karaman \cite{ma2018sparse} & ResNet-50 & $80\times128$ & 8.399 & 0.444 & 0.511 & 0.681 & 0.868 & 5.8 ms & Nvidia GTX 1080 \\
		12 & Ma and Karaman ($5\times5$) \cite{ma2018sparse} & ResNet-50 & $80\times128$ & 4.988 & 0.178 & 0.798 & 0.923 & 0.962 & 6.7 ms & Nvidia GTX 1080 \\
		13 & Ma and Karaman ($3\times3$) \cite{ma2018sparse} & ResNet-50 & $80\times128$ & 4.744 & 0.174 & 0.804 & 0.930 & 0.967 & 6.7 ms & Nvidia GTX 1080 \\         
		14 & Wofk et al. \cite{wofk2019fastdepth} & MobileNet & $80\times128$ & 4.874 & 0.166 & 0.814 & 0.928 & 0.967 & 7.3 ms & Nvidia GTX 1080 \\ 
		15 & MobileXNet (Ours) & $\triangle$ & $80\times128$ & \textcolor{red}{\textbf{3.904}} & 0.140 & 0.853 & 0.948 & 0.975 & 5.8 ms & Nvidia GTX 1080 \\ \hline 		
		
	\end{tabular}
	
	\begin{tabular}{| c | c | c |} \hline
		\cellcolor{gray}Lower is better & \cellcolor{pink}Higher is better & {$\delta_1: \delta < 1.25$}, {$\delta_2: \delta < 1.25^{2}$}, {$\delta_3: \delta < 1.25^{3}$} \\ \hline 
	\end{tabular} 
\end{table*}
\begin{figure*}[h]
	\centering
	\subfloat[]{
		\includegraphics[width=.25\linewidth]{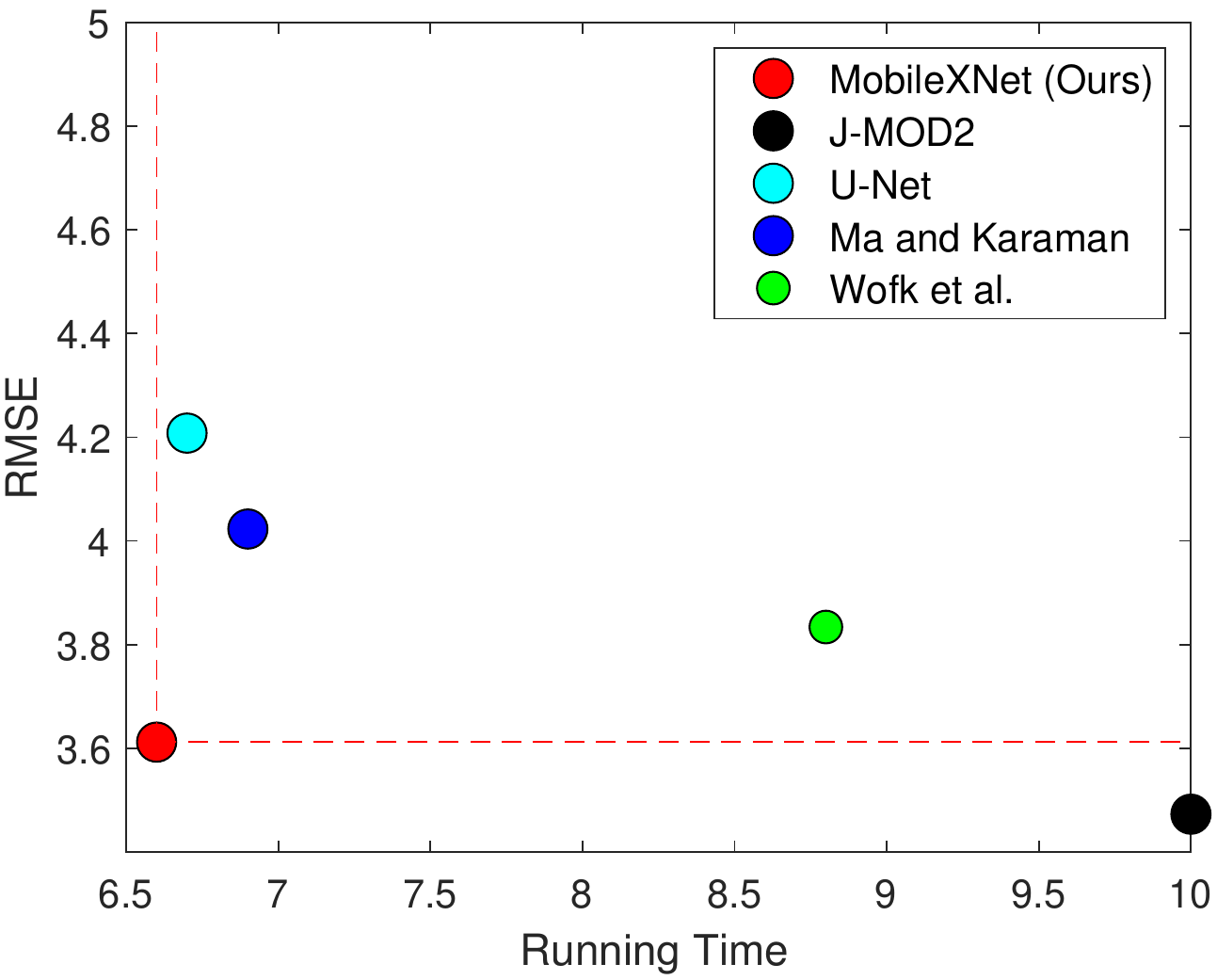}
		\label{fig:original}}
	\hspace{0.16in}
	\subfloat[]{
		\includegraphics[width=.25\linewidth]{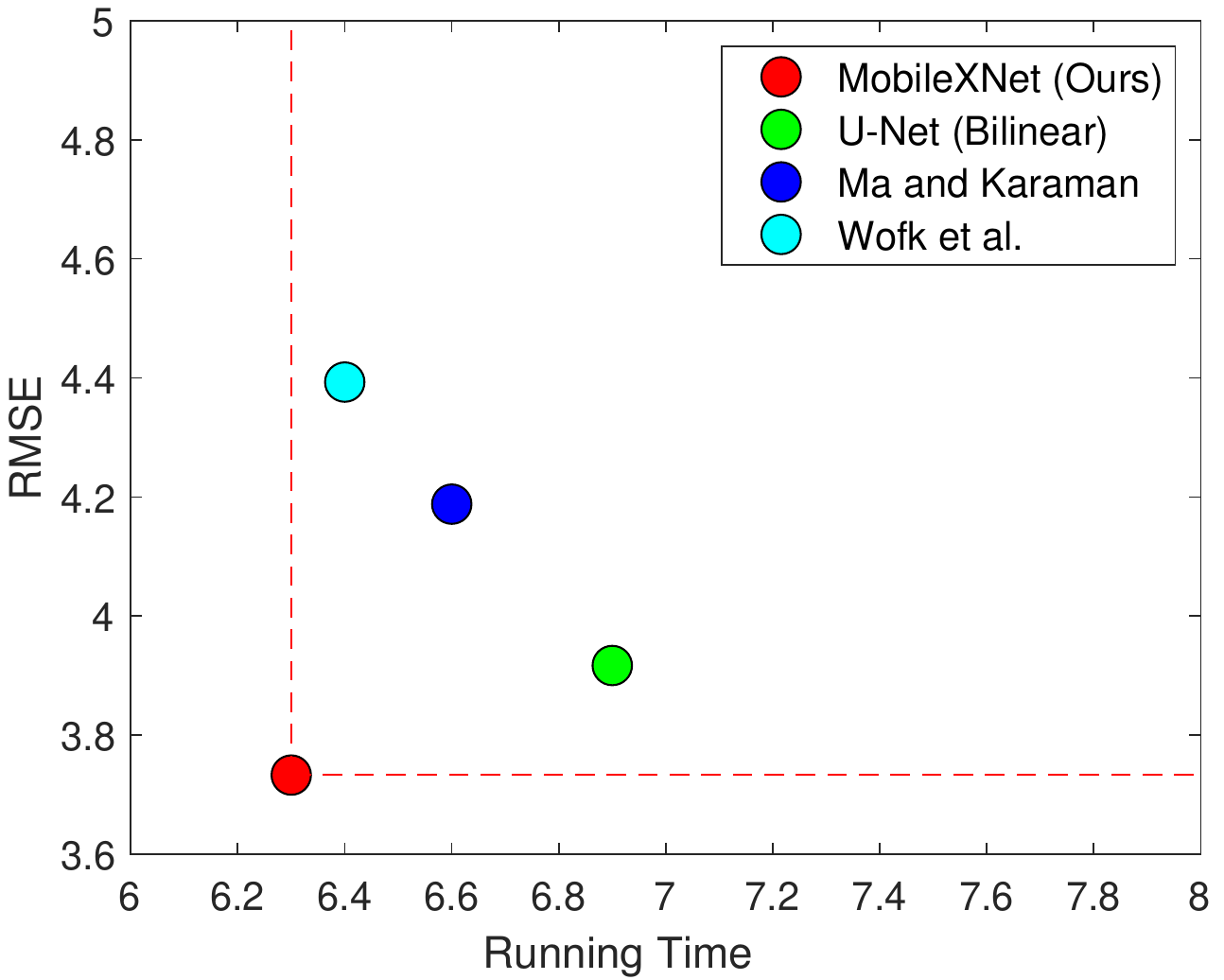}
		\label{fig:128x204}}
	\hspace{0.16in}
	\subfloat[]{
		\includegraphics[width=.25\linewidth]{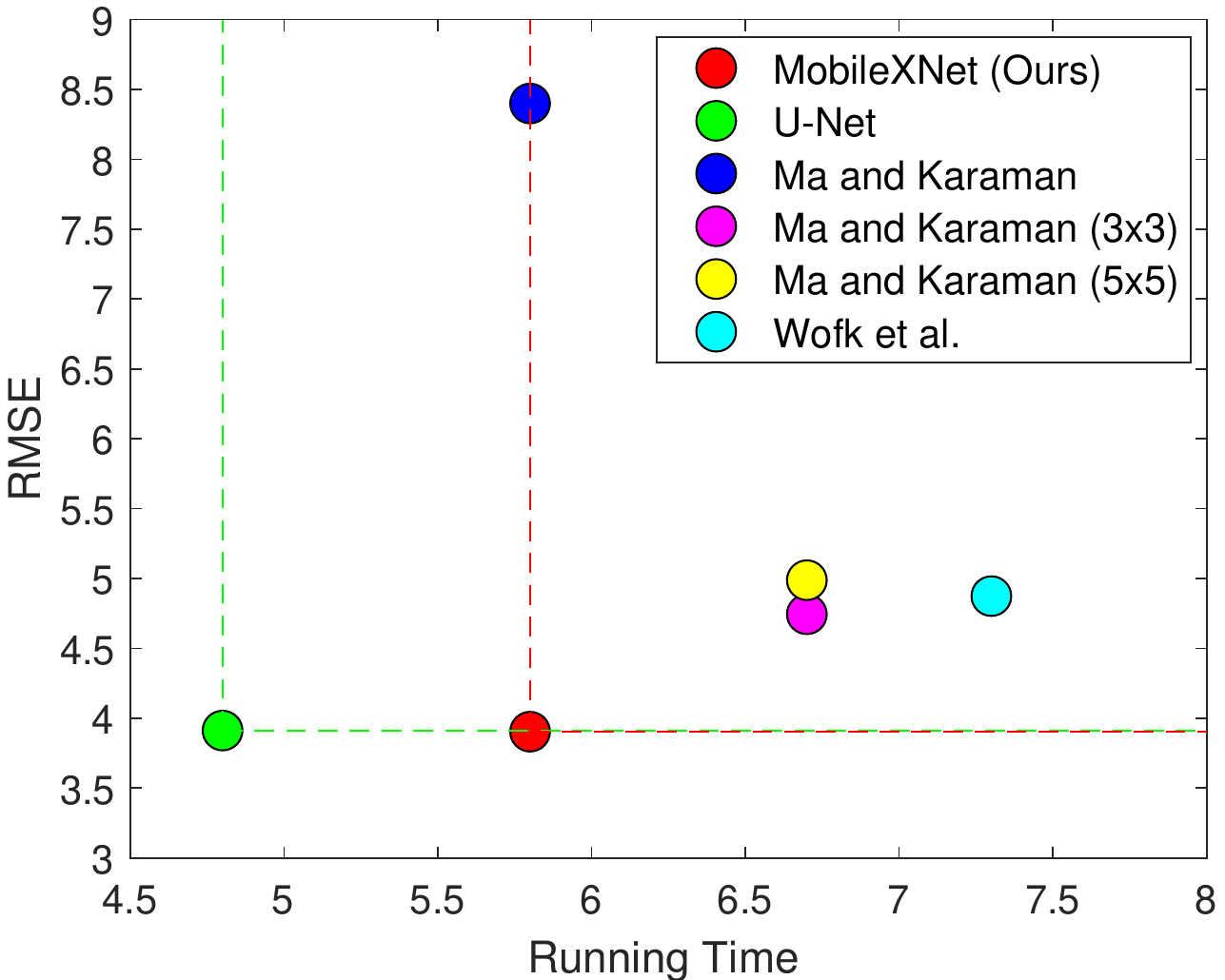}
		\label{fig:80x128}}
	\vspace{-.1in}
	\caption{Pareto Optimality on the UnrealDataset. (a) The original images ($160\times256$), (b) The resized images ($128\times204$), and (c) The resized images ($80\times128$).}
	\label{fig:pareto_rmse_unreal}
	\vspace{-0.1in}
\end{figure*}
To further validate the ability of MobileXNet to process small sized images, we downsample the original images to $128\times204$ and $80\times128$ pixels. We compare our method with \cite{ronneberger2015u, ma2018sparse, wofk2019fastdepth}. The rows 6 to 9 and rows 10 to 15 of Table \ref{table: comparison_on_unreal} show results on $128\times204$ and $80\times128$ sized images respectively. It can be observed that reducing the image size decreases the performance of depth estimation. It is worth noting that when images are downsampled to $80\times128$ pixels, the performance of Ma and Karaman \cite{ma2018sparse} dropped significantly. Ma and Karaman \cite{ma2018sparse} use the fully convolutional part of ResNet-50 as encoder. Besides the inherent characteristic of the encoder network, the $7\times7$ sized filters in the input layer may have influence on the performance. For small sized images, large filters could capture global information, while the detail information are missed. Unlike image classification, depth estimation is a pixel-wise application, both global and detailed information are important. Our hypothesis is that using smaller sized filters in the input layer will help to improve the performance of Ma and Karaman \cite{ma2018sparse}. As a result, we modified \cite{ma2018sparse} by reducing the filter kernel size in the input layer from 7 to 5 and 3 respectively. \par 
It can be observed from rows 11 to 13 of Table VII, both $5\times5$ and $3\times3$ sized filter-based input layers improve the performance of Ma and Karaman \cite{ma2018sparse}. We present a qualitative comparison between our method and baselines in Fig. \ref{fig: qualitive comaprison on the half sized images of Unreal dataset}. It is clear that the modified Ma and Karaman \cite{ma2018sparse} with small sized filter-based input layers performs better than the original. Thus, we can conclude that the input layer with small sized filters is helpful in predicting depth from small sized images. It is worth noting that MobileXNet outperforms the modified Ma and Karaman \cite{ma2018sparse} in all metrics and generates much clearer depth maps, although the latter has a very deep encoder network. Morever, according to Fig. \ref{fig:pareto_rmse_unreal}b and c, the proposed MobileXNet is the non-dominated solution in $128\times204$ and $80\times128$ sized images. This demonstrates the superior performance of MobileXNet on small sized images. \par
\section{Conclusions and Future Work}
In this paper, we paid attention to the trade-off between the accuracy and speed of monocular depth estimation (MDE). To this end, we introduced a novel and real-time CNN architecture, namely, MobileXNet. By assembling two shallow and simple encoder-decoder subnetworks back-to-back in a unified framework, MobileXNet utilizes a simple network architecture and avoids a greater number of successive downsamplings. With the benefit of this design, the proposed method not only achieve a proper trade-off between the accuracy and speed of MDE, but also shows superior performance on the small sized images. Being designed in the encoder-decoder style and trained in the end-to-end manner, the proposed network also can be transferred to other applications, e.g., foreground segmentation \cite{akilan20193d}.

We also designed a hybrid loss function by employing an additional spatial derivative-based $L_{1}$ loss function together with the regular $L_{1}$ function. Comprehensive experiments have been conducted on the NYU depth v2, KITTI, Make3D and Unreal data sets. Experimental results showed that the proposed method performed comparably to the state-of-the-art methods which either were implemented with a complex and extremely deep architecture or used post-processing. More importantly, the proposed MobileXNet run at a much faster speed on a single less powerful GPU, which is particularly critical for real-time autonomous driving and robotic applications.

In the future, we will investigate more light-weight and accurate variants of the MobileXNet network and apply it to obstacle avoidance of MAVs.


\bibliographystyle{IEEEtran}
\bibliography{references}

\end{document}